\pdfoutput=1

\documentclass[11pt]{article}

\usepackage[final]{acl}

\usepackage{times}
\usepackage{latexsym}

\usepackage[T1]{fontenc}

\usepackage[utf8]{inputenc}

\usepackage{microtype}

\usepackage{inconsolata}

\usepackage{graphicx}

\usepackage{amsmath}
\usepackage{amssymb}
\usepackage{mathtools}
\usepackage{amsthm}
\usepackage{amsfonts}
\usepackage{dsfont}

\usepackage[breaklinks]{hyperref}   %
\usepackage[noabbrev,capitalize]{cleveref}
\usepackage{graphicx}
\usepackage{caption}
\usepackage{subcaption}
\usepackage{xspace}
\usepackage[normalem]{ulem}
\usepackage{multirow}
\usepackage{float}
\usepackage{ulem}
\usepackage{multicol}
\usepackage{paralist}   %
\usepackage{enumitem}   %
\usepackage{cuted}
\usepackage{url}
\usepackage{setspace}
\usepackage{fancyhdr}
\usepackage{titlesec}
\usepackage{booktabs}
\usepackage{makecell}
\usepackage{dblfloatfix}
\usepackage{xurl}

\usepackage{amsmath,amsfonts,bm}

\def\eqref#1{equation~\ref{#1}}

\def\1{\bm{1}}

\def\rvh{{\mathbf{h}}}

\def\rvx{{\mathbf{x}}}

\def\rmH{{\mathbf{H}}}

\def\rmW{{\mathbf{W}}}

\DeclareMathAlphabet{\mathsfit}{\encodingdefault}{\sfdefault}{m}{sl}
\SetMathAlphabet{\mathsfit}{bold}{\encodingdefault}{\sfdefault}{bx}{n}

\def\gU{{\mathcal{U}}}

\usepackage{amsmath}

\usepackage[dvipsnames]{xcolor}
\usepackage{xparse}
\usepackage[definitionLists,hashEnumerators,smartEllipses, hybrid]{markdown}

\newcommand{\ifprecedingtext}[1]{\ifvmode\relax\else#1\fi}

\DeclareMathOperator*{\doop}{do}

\newenvironment{redenv}{%
    \color{BrickRed}%
}{%
    \ignorespacesafterend%
}

\newenvironment{blueenv}{%
    \color{blue}%
}{%
    \ignorespacesafterend%
}

\newenvironment{orangeenv}{%
    \color{orange}%
}{%
    \ignorespacesafterend%
}

\newenvironment{purpleenv}{%
    \color{Plum}%
}{%
    \ignorespacesafterend%
}

\newenvironment{oliveenv}{%
    \color{olive}%
}{%
    \ignorespacesafterend%
}

\newenvironment{greenenv}{%
    \color{ForestGreen}%
}{%
    \ignorespacesafterend%
}

\newenvironment{tabenv}
   {\list{}{}%
    \item\relax}
   {\endlist}

\usepackage[many]{tcolorbox}    	%
\usepackage{setspace}               
\usepackage{multicol}               

\definecolor{main}{HTML}{5989cf}    %
\definecolor{sub}{HTML}{cde4ff}     %

\tcbset{
    sharp corners,
    colback = white,
    before skip = 0.2cm,    %
    after skip = 0.5cm      %
}                           %

\newtcolorbox{bbox}{
    colback = sub, 
    colframe = main, 
    boxrule = 0pt, 
    leftrule = 6pt %
}

\newcommand{\mask}{\texttt{[MASK]}\xspace}

\let\oldmax\max
\renewcommand{\max}{complete\xspace}
\newcommand{\maxy}{completeness\xspace}

\newcommand{\Maxy}{Completeness\xspace}
\let\oldmin\min
\renewcommand{\min}{selective\xspace}
\newcommand{\miny}{selectivity\xspace}

\newcommand{\Miny}{Selectivity\xspace}
\newcommand{\cfinlp}{AlterRep\xspace}
\newcommand{\Orc}{Validation Probe\xspace}
\newcommand{\orc}{validation probe\xspace}
\newcommand{\Orcs}{Validation Probes\xspace}
\newcommand{\orcs}{validation probes\xspace}
\newcommand{\Orcps}{Validation probes\xspace}

\renewcommand{\Pr}{P}
\let\Pr\relax
\DeclareMathOperator{\Pr}{P}

\title{How Reliable are Causal Probing Interventions?}

\author{%
  Marc E. Canby$^*$ \quad Adam Davies$^*$ \quad Chirag Rastogi \quad Julia Hockenmaier\\%\thanks{Use footnote for providing further information
  Siebel School of Computing and Data Science\\
  The Grainger College of Engineering\\
  University of Illinois Urbana-Champaign\\
  \texttt{\{marcec2,adavies4,chiragr2,juliahmr\}@illinois.edu} 
}

\begin{document}

\maketitle

\def\thefootnote{*}\footnotetext{These authors contributed equally to this work.}\def\thefootnote{\arabic{footnote}}

\begin{abstract}
Causal probing aims to analyze foundation models by examining how intervening on their representation of various latent properties impacts their outputs. Recent works have cast doubt on the theoretical basis of several leading causal probing methods, but it has been unclear how to systematically evaluate the effectiveness of these methods in practice. 
To address this, we define two key causal probing desiderata: \emph{completeness} (how thoroughly the representation of the target property has been transformed) and \emph{selectivity} (how little non-targeted properties have been impacted). 
We find that there is an inherent tradeoff between the two, which we define as \emph{reliability}, their harmonic mean. 
We introduce an empirical analysis framework to measure and evaluate these quantities, allowing us to make the first direct comparisons between different families of leading causal probing methods (e.g., linear vs. nonlinear, or concept removal vs. counterfactual interventions). We find that: (1) all methods show a clear tradeoff between completeness and selectivity; (2) more complete and reliable methods have a greater impact on LLM behavior; and (3) nonlinear interventions are almost always more reliable than linear interventions.

Our project webpage is available at: \url{https://ahdavies6.github.io/causal_probing_reliability/}

\end{abstract}

\section{Introduction}\label{sec:intro}

What latent properties do large language models (LLMs) learn to represent, and how do they leverage such representations?
Causal probing aims to answer this question by intervening on a model's embedding representations 
of some property of interest (e.g., parts-of-speech), feeding the altered embeddings back into the LLM, and
assessing how the model's behavior on downstream tasks changes %
\cite{geiger2020neural,ravfogel2020inlp,elazar2021amnesic,tucker2021counterfactual,lasri2022probing,davies2023calm,zou2023representation}.
However, it is only possible to draw meaningful conclusions about the model's use of the latent property if we are confident that interventions have fully and precisely carried out the intended transformation \cite{davies2024cognitive}. 
Indeed, prior works have raised serious doubts about causal probing, finding that many intervention methods may have a large unintended impact on non-targeted properties \cite{kumar2022unreliable}, and that the original value of the property may still be recoverable \cite{elazar2021amnesic,ravfogel2022kernel}.
So far, it has been unclear how these doubts generalize to other types of interventions or how serious they are in practice, as there is no generally accepted approach for evaluating or comparing different methods.

\begin{figure*}[t]
    \captionsetup{singlelinecheck=off}
    \centering
    \includegraphics[width=\textwidth]{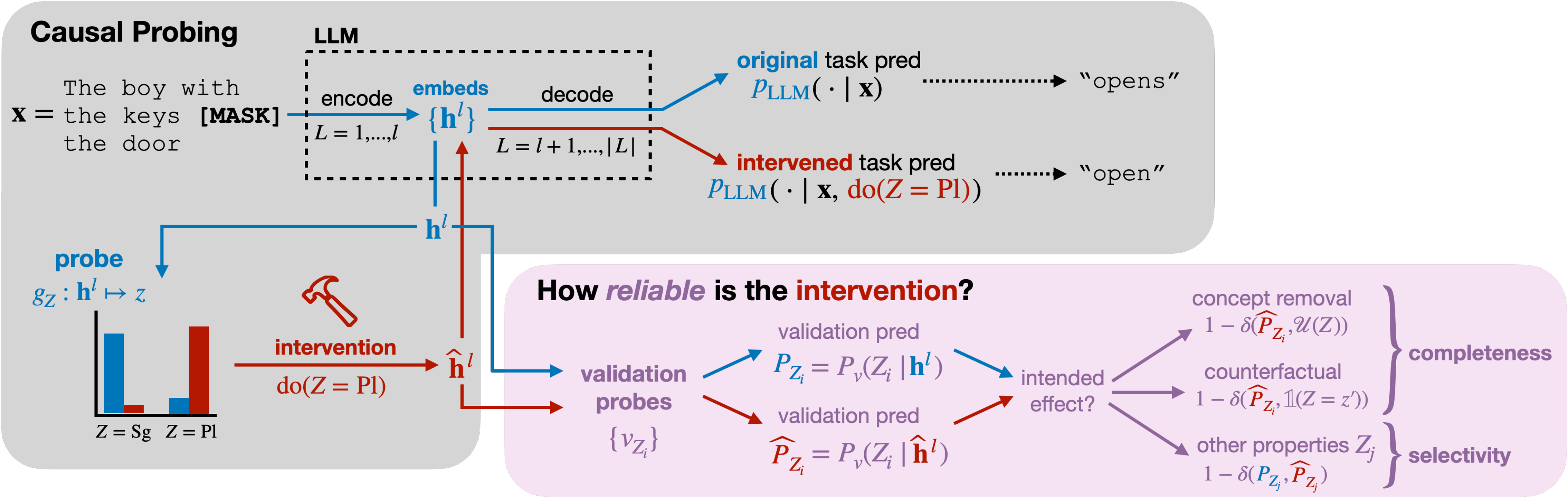}
    \caption[]{\textbf{Causal Probing and Our Reliability Framework.} The process of causal probing is shown in the gray box, with our reliability framework in the purple box.
    \begin{itemize}
        \item \emph{Causal Probing:} embeddings $\rvh^l$ are extracted from layer $L = l$ of a model and used to train a probe $g_Z$ to predict the value $Z = z$ of property $Z$ from embeddings (e.g., the number of the subject, \texttt{boy}, is $Z = \texttt{Sg}$ for singular). A causal probing intervention $\doop(Z = \texttt{Pl})$ uses the probe $g_Z$ to modify the representation encoded by $\rvh^l$ to encode plural instead. The resulting intervened embedding $\hat\rvh^l$ is fed back into the model at layer $L = l+1$ and the forward pass is completed, changing the original prediction \texttt{opens} to the intervened prediction \texttt{open}.
        \item \emph{Reliability Framework:} instead of feeding the intervened embedding $\hat\rvh^l$ back into the model, it is passed alongside $\rvh^l$ to \orcs $\{ v_{Z_i} \}$ that independently test whether the intervention has had the intended effect. \Maxy is measured as the similarity between the \orc prediction and the target distribution for the intervention (e.g., a perfectly \emph{\max} counterfactual intervention $\doop(Z = \texttt{Pl})$ would lead \orc $v_Z$ to predict plural with probability $P_v(Z = \texttt{Pl} | \hat\rvh^l) = 1$), and \miny is the similarity between the \orc distribution for non-targeted properties before and after the intervention (which, for a perfectly \emph{\min} intervention, should not change).
    \end{itemize}
    }
    \label{fig:framework}
\end{figure*}

Thus, our main goal in this study is to work toward a systematic understanding of the effectiveness and limitations of current causal probing methodologies. Specifically, we propose an empirical analysis framework to evaluate the \emph{reliability} of causal probing according to two key desiderata:
\begin{compactenum}
    \item \emph{\Maxy}: interventions should fully transform the representation of targeted properties.
    \item \emph{\Miny}: interventions should not impact non-targeted properties.
\end{compactenum}
We define \maxy and \miny using ``\orcs'' that enable measuring the impact of an intervention on both targeted and non-targeted properties. We apply our framework to several intervention methods and LLMs, observing that each method exhibits a clear tradeoff between these criteria. We also show that
the most complete and reliable interventions lead to the largest and most consistent impact in LLM task performance.
Finally, we find a substantial difference between the reliability of linear versus nonlinear interventions, where nonlinear methods are almost universally more reliable than linear methods across LLMs and between different layers.
This suggests that interventions relying on the linear representation hypothesis (see, e.g., \citealp{vargas2020exploring,ravfogel2020inlp,ravfogel2022adversarial,tigges2023linear,burns2023discovering,jiang2024origins,park2024linear,park2024geometry}) may yield inaccurate interpretations of model internals and behaviors.
Finally, our framework also provides the first concrete basis for calibrating intervention hyperparameters to balance completeness and selectivity, allowing for more reliable interpretation of LLMs using existing methods.

\section{Background and Related Work}\label{sec:relwk}

\paragraph{Probing}\label{sec:sp}
\emph{Probing} 
aims to analyze which properties (e.g., part-of-speech, sentiment labels, etc.) are represented by a deep learning model (e.g., LLM) by training classifiers to predict these properties from latent embeddings  \cite{belinkov2022probing}. 
Given, say, an LLM $M$, input token sequence $\rvx = (x_1, ..., x_N)$, and embeddings $\rvh^l = M_l(\rvx)$ of input $\rvx$ at layer $l$ of $M$, suppose $Z$ is a latent property of interest that takes a discrete value $Z = z$ for input $\rvx$. Here, the formal goal of probing is to train a classifier $g_Z: M_l(\rvx) \mapsto z$ to predict the value of $Z$ from  $\rvh^l$. %
On the most straightforward interpretation, if $g_Z$ achieves high accuracy on the probe task, then the model is said to be ``representing'' $Z$.
An important criticism of such claims is that \emph{correlation does not imply causation} -- i.e., that simply because a given property can be predicted from embedding representations does not mean that the model is using the property in any way \cite{hewitt2019control,elazar2021amnesic,belinkov2022probing,davies2023calm}.

\paragraph{Causal Probing}\label{sec:bgcp}
A prominent response to this concern has been \emph{causal probing}, which uses probes to remove or alter that property in the model's representation, and measuring the impact of such interventions on the model's predictions (\citealp{elazar2021amnesic,tucker2021counterfactual,lasri2022probing,davies2023calm}; see \cref{fig:framework}).
Specifically, causal probing performs interventions $\doop(Z)$ that modify $M$'s representation of $Z$ in the embeddings $\rvh^l$, producing $\hat{\rvh}^l$, where interventions can either encode a counterfactual value $Z = z'$ %
(denoted $\doop(Z = z')$ where $z \neq z'$), or remove the representation of $Z$ entirely (denoted $\doop(Z = 0)$). Following the intervention, modified embeddings $\hat{\rvh}^l$ are fed back into $M$ beginning at layer $l+1$ to complete the forward pass, yielding intervened predictions $\Pr_M( \cdot | \rvx, \doop(Z))$. 
Comparison with the original predictions $\Pr_M(\cdot | \rvx)$ allows one to measure the extent to which $M$ uses its representation of $Z$ in computing them.

\paragraph{Causal Probing: Limitations}\label{sec:bgcplimit}
Prior works have indicated that information about the target property that should have been completely removed may still be recoverable by the model \cite{elazar2021amnesic,ravfogel2022kernel,ravfogel2023linear}, in which case interventions are not \max; or that most of the impact of interventions may actually be the result of collateral damage to correlated, non-targeted properties \cite{kumar2022unreliable}, in which case interventions are not \min.
How seriously should we take such critiques?
We observe several important shortcomings in each of these prior studies on the limitations of causal probing interventions:
\begin{compactenum}
    \item These limitations have only been empirically demonstrated for the  task of removing  information about a target property from embeddings such that the model \emph{cannot be fine-tuned to use the property for downstream tasks} \cite{kumar2022unreliable,ravfogel2022kernel,ravfogel2023linear}. But considering that the goal of causal probing is to interpret the behavior of an existing pre-trained model,
    the question is not whether models \emph{can} be fine-tuned to use the property; it is whether models \emph{already} use the property without task-specific fine-tuning, which has not been addressed in prior work. Do we observe the same limitations in this context?
    \item 
    These limitations have only been studied for linear concept removal interventions (e.g., \citealt{ravfogel2020inlp,ravfogel2022adversarial}),
    despite the recent proliferation of other causal probing methodologies, including nonlinear \cite{tucker2021counterfactual,ravfogel2022kernel,shao2022spectral,davies2023calm} and counterfactual interventions \cite{ravfogel2021counterfactual,tucker2021counterfactual,davies2023calm} (see \cref{sec:int}).
    Do we observe the same limitations for, e.g., nonlinear counterfactual interventions?
\end{compactenum}
In this work, we answer both questions by providing a precise, quantifiable, and sufficiently general definition of completeness and selectivity that it is applicable to \emph{all} such causal probing interventions, and carry out extensive experiments to evaluate 
representative methods from each category of interventions when applied to a pre-trained LLM as it performs a zero-shot prompt task.

\paragraph{Causal Probing: Evaluation}
\label{sec:bgcpeval}
Note that, while we are the first to define and measure the completeness and selectivity of \emph{causal probing interventions},
RAVEL \citep{huang2024ravel} provides a broadly analogous evaluation framework and dataset with respect to \emph{interchange interventions}. %
Methods for performing interchange interventions over embedding representations of a given property are trained on counterfactual minimal pairs of the property (i.e., two inputs which are identical in all respects except the input property; \citealp{geiger2020neural,vig2020causal,geiger2024finding}).
In contrast, causal probing, as studied in this work, does not require minimal pairs for training probes or performing interventions \cite{davies2023calm}, allowing our empirical analysis to be carried out without access to such data.

\section{Evaluating Causal Probing Reliability}\label{sec:eframe}

Recall that our main goal in this work is to evaluate intervention reliability in terms of \maxy (completely transforming $M$'s representation of some target property $Z_i$) and \miny (minimally impacting $M$'s representation of other properties $Z_j \neq Z_i$).\footnote{
    In this paper, we use \miny in the sense described by \citet{elazar2021amnesic}, and not other probing work such as \citet{hewitt2019control}, where it instead refers to the gap in performance between probes trained to predict real properties versus nonsense properties.
}
Given that we cannot directly inspect what value $M$ encodes for any given property $Z_i$, it is necessary to introduce the notion of \textit{\orcs}, which we use to measure the extent to which interventions have fulfilled either criterion. %
Our complete reliability framework is visualized in \cref{fig:framework}.

\paragraph{\Orcs} 
We define a \orc $v$ as a probe (trained independently from interventional probes; see \cref{sec:oprobes}) that returns a
distribution $\Pr_v(Z| \rvh)$ over the values of property $Z$, and we interpret  $\Pr_v(Z = z | \rvh)$ as the 
degree to which the model's embedding representations\footnote{
    For simplicity, we omit the superscript $l$ denoting the layer embeddings $\rvh^l$ from which $\rvh$ is extracted; but our framework can be applied to study interventions over embeddings from any layer.
} $\rvh$ given natural-language input $\rvx$ encodes a belief that $\rvx$ has the property $Z = z$.
So, if $\rvh$ encodes value $Z=\hat z$ with complete certainty, $v$ should return a degenerate distribution $\Pr_v (Z| \rvh) = \mathds{1}(Z=\hat z)$,  whereas we would expect a uniform distribution $\Pr_v(Z| \rvh)=\mathcal{U}(Z)$ if  $\rvh$ does not encode property $Z$ at all.\footnote{
    A \orc's prediction is subtly different from the prediction an arbitrary classifier should make in the absence of any evidence about $Z$: such a classifier should revert to the empirical distribution $\hat\Pr(Z)$.
} 
Thus, \orcs enable us to estimate how well various intervention methods carry out the target transformation. (See \cref{sec:oprobes} for details on \orc training.)

\paragraph{\Maxy}
If a counterfactual intervention $\doop(Z=z')$ is perfectly \emph{\max}, then it would produce a perfectly-intervened $\rvh^*_{Z = z'}$ that fully transforms $\rvh$ from encoding value $Z=z$ to encoding counterfactual value $Z=z' \neq z$. 
Thus, after performing the intervention, \orc $v$ should emit %
$\Pr_v(Z=z'| \rvh^*_{Z = z'}) = P^*_Z(Z = z') = 1$. 
For concept removal interventions  $\doop(Z=0)$, a perfectly complete representation $\rvh^*_{Z=0}$ should not encode $Z$ at all: $\Pr_v(Z| \rvh^*_{Z = 0}) =P^*_Z(Z) = \mathcal{U}(Z)$.\footnote{
    This is only expected when using concept removal interventions for \emph{causal probing} -- i.e., when intervening on a model's representation and feeding it back into the model to observe how the intervention modifies its behavior. When considering concept removal interventions for \emph{concept removal} (a more common setting), a more appropriate ``goal'' distribution $P^*_Z$ would be $\Pr(Z)$, the label distribution. See \cref{sec:null_complete} for further discussion.
}

We can use any distributional distance metric $\delta(\cdot, \cdot)$ bounded by [0, 1] to determine how far the observed distribution $\hat P_Z = \Pr_v(Z| \hat \rvh_{Z})$ is from the ``goal'' distribution $P^*_Z$.
Throughout this work, we use total variation (TV) distance, %
which allows us to directly compare counterfactual and concept removal distributions: in both cases, $0 \leq c(\hat \rvh_{Z}) \leq 1$, where attaining $1$ means the intervention had its intended effect in transforming the encoding of $Z$.
Finally, for a given set of test embeddings $\rmH = \{ \rvh^k \}_{k=1}^n$, the aggregate completeness over this test set $C(\rmH_Z)$ is the average $c(\hat \rvh^i_{Z})$ across all $\rvh^k \in \rmH$.

For \textbf{counterfactual interventions}, we measure completeness as:
\begin{equation}\label{eq:ccf}
c(\hat \rvh_{Z})=1- \delta (\hat P_Z, P^*_Z)
\end{equation}
If the intervention is perfectly complete, then $\hat P_Z = P^*_Z$ and $c(\hat \rvh_{Z})=1$. On the other hand, if $\hat P_Z$ is maximally different from the goal distribution $P^*_Z$ (e.g., $\hat P_Z = \Pr_v(Z=z|   \hat\rvh_{Z=z'})=1$), then $c(\hat \rvh_{Z})=0$.
For properties with more than two possible values, \maxy is computed by averaging over each possible counterfactual value $z'_1, ..., z'_k \neq z$, yielding $c(\hat \rvh_{Z}) = \frac{1}{k}\sum_{i=1}^k\hat c(\rvh_{Z=z'_i})$.

For \textbf{concept removal interventions}, we measure completeness as:
\begin{equation}\label{eq:cn}
c(\hat \rvh_{Z}) = 1-\frac{k}{k-1}\cdot \delta (\hat P_Z, P^*_Z)
\end{equation} 
where $k$ is the number of values $Z$ can take. The normalizing factor is needed because $P^*_Z$ is the uniform distribution over $k$ values and hence $0 \leq \delta (\hat P_Z, P^*_Z) \leq 1-\frac{1}{k}$.

\paragraph{\Miny}
If an intervention on property $Z_i$ 
is \emph{\min}, the intervention should not impact $M$'s  representation of any  non-targeted property $Z_j \neq Z_i$.
Thus, for both counterfactual and concept removal interventions, \orc $v$'s prediction for any such $Z_j$ should not change after the intervention. 

To measure the selectivity of a modified representation $\hat \rvh_{Z_i}$ with respect to $Z_j$, denoted $s_j(\hat \rvh_{Z_i})$, we can again measure the distance between the observed distribution $\hat P_{Z_j} =\Pr_v(Z_j| \hat \rvh_{Z_i})$ and the original (non-intervened) distribution $P_{Z_j}=\Pr_v(Z_j| \rvh)$: \begin{gather}\label{eq:s}
    s_j\big(\hat \rvh_{Z_i}\big)=1 - \frac{1}{m} \cdot \delta \big (\hat P_{Z_j}, P_{Z_j} \big ) \\
    \text{where } m = \oldmax\big(1 - \oldmin(P_{Z_j}), \oldmax(P_{Z_j})\big) \nonumber
\end{gather}
Since $0 \leq \delta(\hat P_{Z_j}, P_{Z_j}) \leq m$, we divide by $m$ to normalize \miny to $0 \leq s_j (\hat \rvh_{Z_i})\leq 1$.
If multiple non-targeted properties $Z_{j_1}, ..., Z_{j_\text{max}}$ are being considered, selectivity $s(\hat \rvh_{Z_i})$ is computed as the average over all such properties $s_{j_m}(\hat \rvh_{Z_i})$.
Finally, analogous to completeness, the aggregate selectivity over a set of test embeddings $\rmH_{Z_i} = \{ \rvh^k \}_{k=1}^n$, denoted $S(\rmH_{Z_i})$, is the average selectivity 
$s(\hat \rvh^k_{Z_i})$
across all $\rvh^k_{Z_i} \in \rmH_{Z_i}$.

\paragraph{Reliability}
Since completeness and selectivity can be seen as a trade-off, we define the overall reliability of an intervention $R(\hat \rmH)$ as the harmonic mean of $C(\hat \rmH ^l)$ and $S(\hat \rmH ^l)$.
This is analogous to the F1-score, which is the harmonic mean of precision and recall: just as a degenerate classifier can achieve perfect recall and low precision by always predicting the positive class, a degenerate intervention can achieve perfect selectivity and low completeness by performing no intervention at all. Using harmonic mean to calculate reliability heavily penalizes such interventions.

\section{Experimental Setting}\label{sec:expsett}

\label{sec:task}

\paragraph{LLMs}
We test our framework in experiments across six language models:
BERT \cite{bert}, GPT-2 \cite{radford2019language}, three Pythia models (160M, 1.4B, and 6.9B; \citealp{biderman2023pythia}), and Llama 3.2 (3B, instruction-tuned; \citealp{grattafiori2024llama3}).
We include BERT and GPT-2 to test causal probing methods in more traditional settings they were originally designed for -- e.g., BERT (an encoder-only masked language model) has been very extensively studied in causal probing \cite{ravfogel2020inlp,rogers2021bertology,elazar2021amnesic,ravfogel2021counterfactual,lasri2022probing,ravfogel2022kernel,ravfogel2023linear,davies2023calm}, and many methods have been designed specifically with this model in mind. 
We include the range of Pythia models to study how these methods scale and generalize to the popular GPT-like family of architectures (decoder-only models trained on autoregressive language modeling). Finally, we account for the effect of popular post-training techniques like instruction-tuning and RLHF by studying the selected Llama model.\footnote{
    For information on post-training of the \texttt{Llama-3.2-3B-Instruct} model used in our experiments, see: \url{https://huggingface.co/meta-llama/Llama-3.2-3B-Instruct\#training-data}
}

\paragraph{Task}
Following several prior causal probing works \cite{lasri2022probing,ravfogel2021counterfactual,arora2024causalgym}, we select the prompting task of \textbf{subject-verb agreement}. (In \cref{apx:ioi}, we also repeat some experiments for the IOI task introduced by \citealt{wang2023ioi}.)
In subject-verb agreement, each data point takes the form $\langle \rvx_i, y_i \rangle$ where $\rvx_i$ is a sentence such as ``the boy with the keys \mask the door,'' and the task of the LLM is to predict 
$\Pr_M(y_i | \rvx) > \Pr_M(y_i' | \rvx)$ (here, that $y_i =$ ``locks'' rather than $y_i' =$ ``lock'') -- see the example in \cref{fig:framework}.
The causal variable $Z_c$ is the number of the subject (\texttt{Sg} or \texttt{Pl}), because (grammatically) this is the only variable that determines the number of the verb in English. The environmental (non-causal) variable $Z_e$ is the number (\texttt{Sg} or \texttt{Pl}) of the noun immediately preceding the verb to be conjugated when that noun is not the subject
(e.g., ``keys'' in the phrase ``with the keys'').
Note that, in this work, we consider a task in the simplest experimental setting (two binary properties) that allows us to study interventions using our framework; however, nothing in our methodology precludes the use of more properties, or properties with more possible values.

\paragraph{Dataset}
We use the LGD dataset \cite{linzen_assessing_2016}, which consists of $>$1M naturalistic English sentences from Wikipedia; from this we take only sentences for which both singular and plural forms of the target verb are in LLMs' vocabularies. We use 40\% of the examples to train \orcs, 40\% to train interventional probes, and 20\% as a test set. (More dataset details can be found in Appendix \ref{sec:dataset}.)

\paragraph{\Orcs}
\label{sec:oprobes}
For each layer $l$ and probed property $Z$, we experiment with several instantiations of \orcs, including linear and MLP probes across a range of hyperparameters (\cref{apx:probe}), observing similar results between them (see \cref{sec:linorc}). 
Thus, for all results reported in the main paper, we default to the \orc architecture and hyperparameters with the highest validation-set accuracy for the probed property.\footnote{
    Note that this always results in MLP \orcs; see \cref{sec:linorc} for results with linear \orcs.
    Validation sets used for selecting \orcs have the same independence and label-distribution properties as their train sets.
}
\Orcps are trained on data that is completely disjoint from that used to train interventional probes, and where $Z_c$ and $Z_e$ are made independent by subsampling the largest (random) subset that preserves label distributions $\Pr(Z_c), \Pr(Z_e)$. This is important for \orcs to serve as unbiased arbiters of selectivity, as spurious correlations between the variables could lead a probe that is trained on property $Z_c$ to partially rely on representations of $Z_e$ \cite{kumar2022unreliable}.\footnote{
    We leave these spurious correlations in training data of interventional probes to test the impact that they have on the resulting interventions' completeness and selectivity, as this is a better proxy for their completeness and selectivity in more realistic settings where controlling for spurious correlations may not always be possible.
}

\begin{figure*}[ht!]
    \centering
    \begin{subfigure}[t]{0.48\textwidth}
        \centering
        \includegraphics[width=\linewidth]{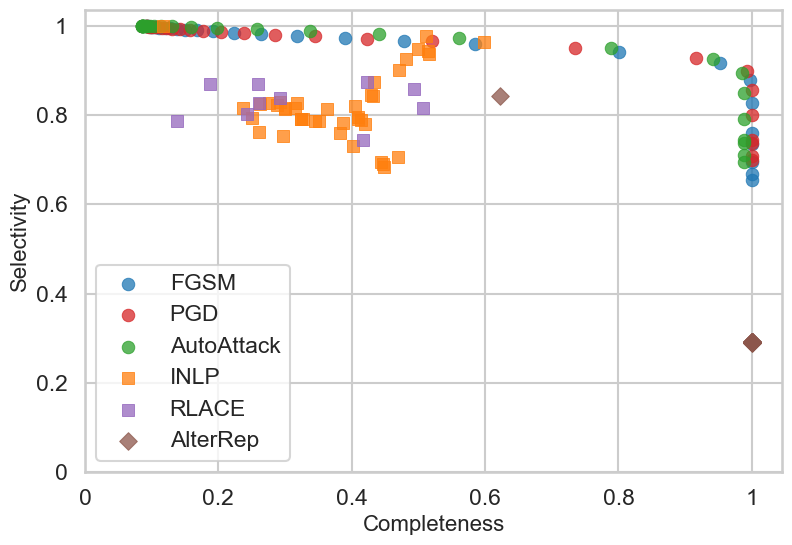}
        \caption{\textbf{Selectivity vs. Completeness}}
        \label{fig:comp_sel_pythia}
    \end{subfigure}
    \hfill
    \begin{subfigure}[t]{0.48\textwidth}
        \centering
        \includegraphics[width=\linewidth]{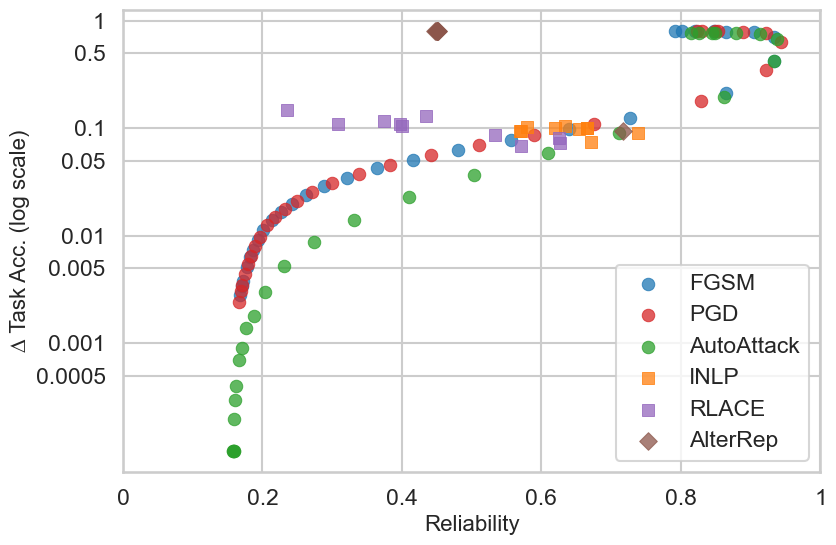}
        \caption{\textbf{$\Delta \text{Task Acc}$ vs. Reliability}}
        \label{fig:acc_rel_pythia}
    \end{subfigure}
    \caption{\textbf{Completeness, selectivity, reliability, and $\Delta \text{Task Acc}$} for all interventions in the \textbf{final layer of Pythia-160M.} Each point in both plots corresponds to a different hyperparameter setting. (\cref{sec:interhyp} contains analogous results for all other models.)}
    \label{fig:main_plots_pythia}
\end{figure*}

\paragraph{Interventions}\label{sec:int}
We explore two (linear) concept removal interventions: INLP \cite{ravfogel2020inlp}, which iteratively trains classifiers on $Z$ and projects embeddings into their nullspaces; and RLACE \cite{ravfogel2022adversarial}, which identifies a minimal-rank subspace to remove information that is linearly predictive of $Z$ by solving a constrained minimax game.
We explore one linear counterfactual method, AlterRep \cite{ravfogel2021counterfactual}, which builds on INLP by projecting embeddings along classifiers' rowspaces, placing them on the counterfactual side of the separating hyperplanes. Finally, we study three nonlinear counterfactual methods, which are all gradient-based interventions \cite{davies2023calm}:
a MLP probe is trained on $Z$, then gradient-based (``white-box'') adversarial attacks are applied to minimize the loss of the probe with respect to the target counterfactual value $Z = z'$ within an $L_\infty$-ball of radius $\varepsilon$ around the original embedding. We experiment with three gradient attack methods -- FGSM \cite{fgsm}, PGD \cite{pgdattack}, and AutoAttack \cite{autoattack} -- as described in \cref{apx:interventions}.
After intervening on $Z_c$ to obtain representations $\hat \rvh_{Z_c}$, we %
use \orcs $\hat o$ %
to measure completeness,
selectivity, and reliability.
Due to compute limitations, we restrict our analysis for Pythia-1.4B and -6.9B to three of the six methods: INLP, AlterRep, and FGSM. Note that this includes at least one method from each of the three classes of methods defined above (linear removal, linear counterfactual, and nonlinear counterfactual, respectively).

\paragraph{Impact on Model Behavior} The ultimate goal of causal probing is to measure a model $M$'s use of a property $Z$ by comparing intervened predictions $\Pr_M(\cdot |  \rvx, \doop(Z))$ to its original predictions $\Pr_M(\cdot |  \rvx)$. Our framework aims to measure the reliability of the interventions themselves, a prerequisite to making claims about the underlying model. 
It is nonetheless important to consider how the completeness, selectivity, and reliability of a given intervention relate to its impact on model behavior.
Thus, for each intervention, we also feed intervened final-layer embeddings $\hat \rvh^L$ for all test instances back into models immediately before word prediction, measuring task accuracy based on whether they assign the correct verb form a higher probability, and subtract this ``intervened'' accuracy from the original task accuracy ($98.62$\%) for each intervention to yield $\Delta \text{Task Acc}$ %
(cf. \citealp{elazar2021amnesic,lasri2022probing,davies2023calm}). %

\section{Experimental Results}
Below, we present results for \maxy, \miny, reliability, and $\Delta \text{Task Acc}$ of all intervention methods in models' final layer (\cref{sec:finallayerresults}),
then examine their reliability in earlier layers (\cref{sec:earlierlayerresults}). Note that, while we only have space to include plots for Pythia-160M (henceforth referred to as ``Pythia'') in this section of the main paper, analogous plots for the other models are available in \cref{sec:all_supp_results}.

\subsection{Final-Layer Results}\label{sec:finallayerresults}

\begin{table*}[t!]
\centering
\resizebox{0.75\textwidth}{!}{%
\begin{tabular}{l|cccccc}
\toprule
           & BERT           & GPT2          & Pythia-160M        & Pythia-1.4B   & Pythia-6.9B & Llama-3.2-3B    \\ \midrule    %
INLP       & 0.464          & 0.106         & 0.739          & 0.841         & 0.751          & 0.668 \\
RLACE      & 0.429          & 0.495         & 0.627          &               &                \\ \midrule
AlterRep   & \textbf{0.835} & 0.427         & 0.717          & 0.597         & 0.519          & 0.891 \\ \midrule
FGSM       & 0.552          & 0.958         & 0.934          & \textbf{0.920} & \textbf{0.951} & \textbf{0.960} \\
PGD        & 0.509          & \textbf{0.98} & \textbf{0.943} &               &                \\
AutoAttack & 0.514          & 0.97          & 0.938          &               &                \\ \bottomrule
\end{tabular}}
\caption{\textbf{Intervention scores for maximum-reliability hyperparameters} in the \textbf{final layer} of all models. (See \cref{sec:earlierlayerresults} for results in earlier layers.) Scores are reported
for the hyperparameter $x_{opt}$ that maximizes the reliability of each respective method, and the highest-reliability method for each model is bolded.
}
\label{tab:alpha}
\end{table*}

First, we note that both \orcs are able to consistently predict each property (97.3\% and 94.4\% accuracy for $Z_c$ and $Z_e$, respectively),
which is a necessary prerequisite to validate any further results.

\paragraph{Completeness, Selectivity, \& Reliability} Each intervention has a hyperparameter ($\varepsilon$ for GBIs, rank $r$ for INLP and RLACE, and $\alpha$ for AlterRep), where increasing its value leads to stronger interventions. Thus, each hyperparameter setting yields a different value of completeness, selectivity, and reliability for a given intervention. 
Figure \ref{fig:comp_sel_pythia} plots selectivity against completeness for each method in Pythia's final layer, showing that increasing the hyperparameter values yields higher completeness and lower selectivity. (Analogous results for other models, as well as plots of completeness, selectivity, and reliability broken down by method and hyperparameter value, are available in \cref{sec:interhyp}.)

Table \ref{tab:alpha} shows these metrics for each method at the hyperparameter that yields the highest reliability.
For all models other than BERT, the nonlinear counterfactual methods (GBIs: FGSM, PGD, and AutoAttack) have the highest overall reliability; for BERT only, \cfinlp is most reliable; and otherwise, the linear methods (both removal and counterfactual) tend to show middling reliability, varying from a low of 0.106 for INLP on GPT2 to a high of 0.841 for INLP on Pythia-1.4B.

\paragraph{Task Accuracy}
Figure 
\ref{fig:acc_rel_pythia} shows $\Delta \text{Task Acc}$ as a function of the reliability for each intervention and hyperparameter setting.
For most methods and hyperparameter values, $\Delta \text{Task Acc}$ increases alongside %
intervention reliability.
Notably, the points at which the GBIs (FGSM, PGD, and AutoAttack) achieve the highest $\Delta \text{Task Acc}$ are \textit{not} at their highest reliability values, resulting in a backward curve %
visible at the top of Figure \ref{fig:acc_rel_pythia}, corresponding to hyperparameter $\varepsilon$ being raised past the point of maximum reliability where there is near-perfect completeness %
but much lower selectivity (see \cref{sec:hyper_variation}).
Finally, RLACE and INLP shows a similar impact on task accuracy even for different completeness and reliability scores due to its ``noisy'' equilibrium in reliability and completeness for high rank $r$ (see \cref{sec:hyper_variation}).

\subsection{Reliability by Layer}\label{sec:earlierlayerresults}
\begin{figure}
    \centering
    \includegraphics[width=0.95\columnwidth]{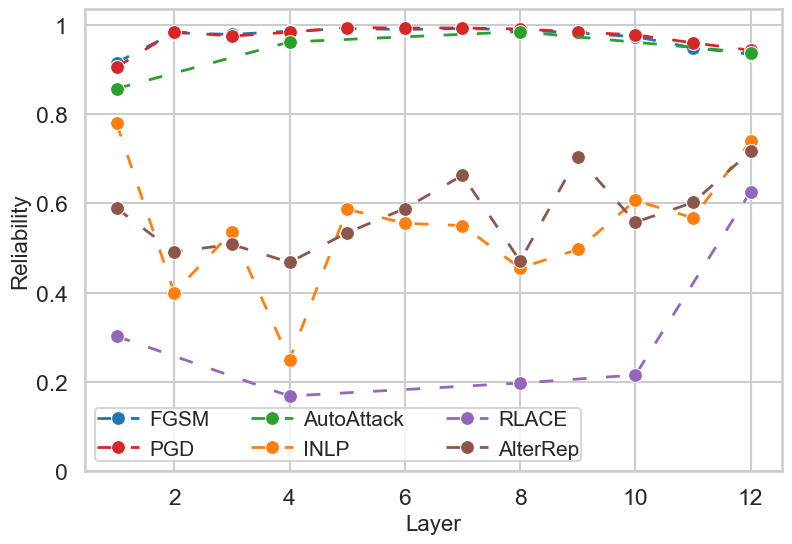}
    \caption{\textbf{Maximum reliability by layer} for each intervention across \textbf{all layers of Pythia-160M.}
    (\cref{sec:layerwise_other_models} contains analogous results for all other models.)
    }
    \label{fig:layerwise_pythia}
\end{figure}

As in the final layer, \orcs over earlier layers can consistently predict each property with high accuracy (see \cref{apx:layerwise_probe}).
\cref{fig:layerwise_pythia} shows the reliability for each method using the hyperparameters that obtain the highest reliability in that layer. Across all layers, each nonlinear counterfactual method (GBIs: FGSM, PGD, and AutoAttack) is more reliable than all linear methods. We also observe this trend for all other models \cref{sec:layerwise_other_models} (with the exception of BERT, for which \cfinlp is more reliable than the GBIs in layers 10-12; see \cref{fig:layerwise_all_others}).

\section{Discussion}

\paragraph{Tradeoff: \Maxy vs. \Miny} 
In Pythia's final layer, no method is able to achieve perfect completeness without sacrificing selectivity (see \cref{fig:comp_sel_pythia}), a trend which we also see for all other models (see \cref{sec:interhyp}). However, we observe a very favorable tradeoff for GBIs, which incur only a small selectivity cost for increasing completeness, leading to high overall reliability across all models and layers (see \cref{sec:layerwise_other_models}). In contrast, linear removal methods (INLP and RLACE) tend to have much higher selectivity than they do completeness, which is likely because these methods are explicitly optimized to minimize collateral damage \cite{ravfogel2020inlp,ravfogel2022adversarial}, despite being linear and potentially incomplete (with respect to putatively nonlinear representations). Finally, the linear counterfactual method (\cfinlp) tends to have highly variable behavior between different models: for Pythia-160M, Llama 3.2, and GPT2, it shows middling performance across layers; it is the \emph{most} reliable method in BERT's last 3 layers, with high selectivity and near-perfect completeness; and it is the \emph{least} reliable method in the later layers of Pythia-1.4B and Pythia-6.9B.

\paragraph{Reliability and Task Accuracy}
\cref{fig:acc_rel_pythia,fig:acc_rel_bert,fig:acc_rel_gpt2} show a clear trend: more complete interventions and hyperparameter values show a greater impact on task performance.
In particular, counterfactual methods (GBIs and \cfinlp), which show consistently higher completeness than removal methods (INLP and RLACE), also show (near-)total $\Delta \text{Task Acc}$.
This is highly intuitive: in the case where models perform the subject-verb agreement task by leveraging its representation of $Z_c$, then more complete interventions %
would have a greater effect on the model's task performance.
We do not claim that this is necessarily the case -- e.g., our results might have looked different if we 
had intervened in earlier layers; and the primary object of our study is the completeness, selectivity, and reliability of the causal probing methods we have experimented with, not the representations used by LLMs to perform a simple grammatical task. Rather, we take the clear relationship between intervention completeness and $\Delta \text{Task Acc}$ to be a strong indicator that more complete methods indeed yield stronger results, reinforcing the utility of our framework in evaluating causal probing interventions as tools for studying models' use of latent representations.
In particular, our framework provides the first concrete approach for calibrating intervention hyperparameters in the latent space (i.e., max-reliability hyperparameter search using validation probes), allowing researchers to adaptively balance the priorities of completeness and selectivity and examine the corresponding effect on model behaviors, rather than simply resorting to maximum-strength \cite{tucker2021counterfactual} or minimum-collateral damage \cite{ravfogel2020inlp,ravfogel2022adversarial} interventions.

\paragraph{Linearity by Layer}
Overall, the nonlinear GBI methods are more reliable than the linear methods across all models and layers, (with the sole exception of BERT in layers 10-12; see \cref{sec:layerwise_other_models}). Without adequate controls, this might simply be the result of using MLP \orcs in all results reported in the main paper, which could bias our analysis in favor of nonlinear methods, as \orcs and nonlinear interventional probes might be relying on similarly-encoded information and neglecting linearly-encoded information. We account for this possibility by repeating layerwise reliability experiments using linear \orcs in \cref{sec:linorc}, finding that they show remarkably similar results to MLP \orcs.

Thus, we briefly consider the more interesting possibility that the reliability gap between linear and nonlinear LLMs \emph{may be due to LLMs encoding task-relevant representations nonlinearly}, particularly in intermediate layers:
for instance, in addition to the aforementioned example of BERT's last 3 layers, Pythia-160M also shows that all linear methods are substantially more reliable in the first and last layers than they are in intermediate layers. While this conjecture is not fully supported by all results (e.g., INLP and \cfinlp drop substantially in reliability in GPT2's final layer), 
it is nonetheless intuitive that some models may be more nonlinear in intermediate layers than their final layer, as embeddings in earlier layers will pass through many nonlinearities before word prediction, allowing a high degree of nonlinear representation \cite{white2021nonlinear}; whereas any output-discriminative information must be made linearly separable in the final embedding layer of neural networks \cite{alain2017understanding}.
There is a long history of work studying the so-called \emph{linear representation hypothesis} (LRH; \citealp{mikolov2013linguistic,pennington2014glove,bolukbasi2016man,vargas2020exploring}) -- i.e., that neural networks encode most or all features linearly -- with some recent works suggesting that this hypothesis is true even for modern LLMs \cite{burns2023discovering,tigges2023linear,park2024linear}.
However, many of these studies often consider embeddings only in the input or final (``unembedding'') layer of LLMs \cite{jiang2024origins,park2024linear,park2024geometry}, neglecting intermediate layers.
Our findings provide an important contrast: while they do not directly validate or refute the LRH, 
the stark difference between the reliability of linear and nonlinear counterfactual methods indicates that it is critical to consider multiple layers throughout models when studying the LRH, as findings of linearity in the final layer may not generalize to earlier layers.

\section{Conclusion}
In this work, we proposed a general empirical evaluation framework for causal probing, defining the reliability of %
interventions in terms of completeness, selectivity, and reliability.
Our framework makes it possible to directly compare different kinds of interventions, such as linear vs. nonlinear or counterfactual vs concept removal methods.
We applied our framework to study leading causal probing techniques across a range of LLMs, finding that they all exhibit a tradeoff between completeness and selectivity, that more reliable and complete methods yield a greater impact on LLM task performance, and that nonlinear methods tend to be much more reliable than linear methods. Finally, we explored the implications of these findings for future work in optimizing intervention hyperparameters and studying the linear representation hypothesis.

\section*{Limitations}
An important empirical limitation of our work is that we only study the relatively simple subject-verb agreement task (and IOI; see \cref{apx:ioi}). We intentionally select simple, well-studied syntactic tasks with a single binary causal variable and one binary environmental variable, opting for a more parsimonious task in this setting to avoid introducing exogenous confounds while studying a novel latent-space evaluation framework in the context of several highly distinct families of methods. Selecting simple tasks also allows for easy comparison between a range of LLMs at different scales (which are all able to solve the task nearly perfectly). However, now that we have validated our framework in the context of these simple tasks, it will be important to extend this study to more complex and interesting tasks, such as those with multiple causal variables that take an arbitrary number of possible values, or those that even frontier models struggle to solve (for use in ``debugging'' what representations are being learned and used by LLMs in performing difficult tasks). We aim to explore such settings in future work.

\section*{Acknowledgements}

This research used the Delta advanced computing and data resource which is supported by the National Science Foundation (award OAC 2005572) and the State of Illinois. Delta is a joint effort of the University of Illinois Urbana-Champaign and its National Center for Supercomputing Applications. This work also utilizes resources supported by the National Science Foundation’s Major Research Instrumentation program, grant \#1725729, as well as the University of Illinois at Urbana-Champaign.
Adam Davies is supported by the National Science Foundation and the Institute of Education Sciences, U.S. Department of Education, through Award \#2229612 (National AI Institute for Inclusive Intelligent Technologies for Education). Any opinions, findings, and conclusions or recommendations expressed in this material are those of the author(s) and do not necessarily reflect the views of National Science Foundation or the U.S. Department of Education.

\bibliography{__citations}

\appendix
\onecolumn

\section{Framework Details}

\paragraph{Completeness of Concept Removal Interventions}\label{sec:null_complete}
In \cref{eq:cn}, we define the ``goal'' distribution $P^*_Z$ of a concept removal intervention used in causal probing as being the uniform distribution -- i.e., for a perfect concept removal intervention, $P^*_Z = \Pr_v (Z| \rvh^*_{Z=0})=\gU(Z)$.
However, this is only true in the case of \emph{causal probing}, and is not true of some concept removal applications such as guarding protected attributes (see, e.g., \citealt{ravfogel2023linear}).
That is, in the case of causal probing, the goal of an intervention is to intervene on a model's representation during its forward pass, feeding the intervened embedding back into the model and observing the change in the model's behavior (as described in \cref{sec:bgcp}).
Recall that the purpose of a \orc $v$ is to decode model $M$'s representation of a given property $Z$, not to predict its ground truth value -- that is, even if $M$ encodes the incorrect value of $Z = z'$ rather than $Z = z$ for a given input, the \orc should still decode the incorrect value $Z = z'$. Indeed, this is precisely the principle behind using \orcs in the case of counterfactual interventions that change the representation of $Z = z$ to counterfactual value $Z = z'$, where \orcs are used to validate the extent to which the representation has actually been changed to encode this counterfactual value, and the ideal counterfactual intervention yields $\Pr_v(Z = z' |  \rvh^*_{Z=z'}) = P^*(Z = z') = 1$.
However, in the case of concept removal interventions $\doop(Z = 0)$, 
an intervened embedding $\rvh^*_{Z=0}$ would ideally remove all information encoding $M$'s representation of the value taken by $Z$, meaning that the $M$ would not encode any value $Z = z_1$ as being more probable than $Z = z_2$ (as any information that is predictive of the value taken by $Z$ should have been removed).
In this case, the \orc $v$ would predict an equal probability $\Pr_v (Z = z_i | \rvh^*_{Z=0})$ for any given value $z_i$ that may be taken by $Z_i$ -- i.e., $\Pr_v (Z | \rvh^*_{Z=0}) = P^*_Z = \gU(Z)$.

However, this is not the case in the context of instances such as guarding protected attributes, where the goal of an intervention $\doop(Z = 0)$ is to remove all information that is predictive of $Z$ from embedding representations $\rvh^*_{Z=0}$ \emph{such that no probe} $g$ \emph{can be trained to predict} $\Pr_g (Z |  \rvh^*_{Z=0})$ any better than predicting $\Pr (Z)$ -- i.e., ignoring the embedding entirely and simply mapping every input to the label distribution $\Pr(Z)$ \citep{ravfogel2023linear}.
In this case, the probe $g$ is trained on intervened embeddings $\hat\rvh_{Z=0}$, in which case it can learn to map every such embedding to the label distribution $\Pr(Z)$, which yields superior performance relative to predicting the uniform distribution $\gU(Z)$ in any case where the label distribution $\Pr(Z)$ is not perfectly uniform, as such a $g$ would have an expected accuracy equal to the proportion of test instances with the most common label $Z = z_{\text{argmax}}$ (which would be greater than the accuracy $\frac{1}{k}$ expected by defaulting to $\gU(Z)$).

The key technical distinction between these two use cases of concept removal interventions is \emph{whether or not probes or underlying models are trained or fine-tuned in the context of interventions}.
In the case of causal probing, they are not -- the (frozen) model $M$ has no opportunity to recover the original value of $Z = z$ following a concept removal intervention $\doop(Z = 0)$, and this should be reflected by \orcs.
This is natural, given that the purpose of causal probing is to interpret the properties used by $M$ in making a given prediction, not to test whether $M$ can be trained to recover properties removed by interventions; and this is reflected by \orcs $v$, which are never trained on intervened embeddings.
In contrast, for concept removal, probes (or models) are trained on intervened embeddings, and may learn to recover properties removed by interventions, meaning that -- even in the worst case where all information has been removed -- it would at least be possible to learn to reproduce the label distribution $\Pr(Z)$; but there is no reason to expect a model $M$ or \orc $v$ to do so, given that they have never been trained on intervened embeddings.
Thus, while we define the ``goal'' distribution $P^*_Z = \gU(Z)$ for measuring the completeness of concept removal interventions as being $\gU(Z)$ rather than $\Pr(Z)$, this distribution would instead be $P^*_Z = \Pr(Z)$ in the case of concept removal.

\section{Experimental Details}

\subsection{LGD Dataset} 
\label{sec:dataset}
We use syntax annotations to extract values for the environmental variable $Z_e$ from the LGD dataset \cite{linzen_assessing_2016}: if the part-of-speech of the word immediately preceding the \mask token is a noun, and it is the object of a preposition (i.e., not the subject), then its number defines $Z_e$. About 83\% of the sentences do not have a prepositional object preceding \mask, and so are only relevant for causal interventions.

The contingency table for values of $Z_c$ and $Z_e$ in the test set are in Table \ref{tab:contingency}.

\begin{table}
\small
\centering
\begin{tabular}{c | ccc | c}
\toprule
    & $Z_e=\texttt{Ø}$  & $Z_e=\texttt{Sg}$  & $Z_e=\texttt{Pl}$  & Total \\ \midrule
$Z_c=\texttt{Sg}$   & 176K & 31K & 5K & 213K \\
$Z_c=\texttt{Pl}$    & 78K & 10K & 4K & 92K \\ \midrule
Total & 254K & 41K & 9K & 305K \\ 
\bottomrule
\end{tabular}
\caption{\textbf{Contingency Table on Test Set}. Distribution of data across combinations of causal and environmental variables. $Z_e=\texttt{Ø}$ denotes instances which have no prepositional phrase attached to the subject (and thus, contain no environmental variable).
Note that the label distributions are unbalanced: $\Pr(Z_c=\texttt{Sg})=69.8\%$ and $\Pr(Z_e=\texttt{Sg}| E \neq \texttt{Ø})=81.5\%$.
}
\label{tab:contingency}
\end{table}

\subsection{Probe Details}\label{apx:probe}

Our experiments include linear and MLP probes (both for interventions and as \orcs). Linear interventions (INLP, RLACE, and \cfinlp) require linear probes; and for nonlinear interventions (GBIs), we use MLPs. We implement probes using PyTorch \cite{paszke2019pytorch}, and leverage LLM implementations of all models available via HuggingFace Transformers \cite{transformerslib}.
For \orcs, we experiment with both linear and MLP probes.
For all probes, we select hyperparameters by performing a grid search across candidate hyperparameter values, selecting the hyperparameters that yield the highest validation-set accuracy.
We save probe parameters from the epoch with the highest validation-set accuracy with patience of 4 epochs.
All probes are trained with cross-entropy loss.

For all \textbf{linear probes}, we consider learning rates in [0.0001, 0.001, 0.01, 0.1].

For \textbf{MLP probes}, we perform grid search over the following hyperparameter values:
\begin{itemize}[noitemsep]
    \item Number of hidden layers: [1, 2, 3]
    \item Layer size: [64, 256, 512, 1024]
    \item Learning rate: [0.0001, 0.001, 0.01]
\end{itemize}

\subsection{Interventions}\label{apx:interventions}
\label{apx:int_ctfl}

\paragraph{Gradient Based Interventions}
For all gradient-based intervention methods \cite{davies2023calm}, we define the maximum perturbation magnitude of each intervention as $\varepsilon$ (i.e., $|| \hat \rvh_Z - \rvh ||_\infty \leq \varepsilon$), and experiment over a range of $\varepsilon$ values between $0.005$ to $5.0$ -- specifically, $\varepsilon \in [0.005$, $0.006$, $0.007$, $0.009$, $0.011$, $0.013$, $0.016$, $0.019$, $0.024$, $0.029$, $0.035$, $0.042$, $0.051$, $0.062$, $0.07$6, $0.092$, $0.112$, $0.136$, $0.165$, $0.2$, $0.286$, $0.409$, $0.585$, $0.836$, $1.196$, $1.71$, $2.445$, $3.497$, $5.0$]. 
We consider the following gradient attack methods for GBIs:
\begin{compactenum}
    \item \textbf{FGSM}
    We implement Fast Gradient Sign Method (FGSM; \citealp{fgsm}) interventions as:
    $$h' = h + \varepsilon \cdot \text{sgn}\left(\nabla_h \mathcal{L}\left(f_{\text{cls}}, x, y\right)\right)$$
    
    \item \textbf{PGD}
    We implement Projected Gradient Descent (PGD; \citealp{pgdattack}) interventions as $ h' = h^T $ where
    $$h_{t+1} = \Pi_{\mathcal{N}(h)} \left( h_t + \alpha \cdot \text{sgn}\left(\nabla_h \mathcal{L}(f_{\text{cls}}, x, y)\right) \right)$$
    for iterations \( t = 0, 1, \ldots, T \), projection operator \( \Pi \), and \( L_\infty \)-neighborhood \( \mathcal{N}(h) = \{ h' : \|h - h'\| \leq \varepsilon \} \).
    For PGD, we use 2 additional hyperparameters: iterations $T$ and step size $\alpha$, while fixing $T = 40$, as suggested by \cite{davies2023calm}.

    \item \textbf{AutoAttack}
    AutoAttack \cite{autoattack} is an ensemble of adversarial attacks that includes FAB, Square, and APGD attacks. Auto-PGD (APGD) is a variant of PGD that automatically adjusts the step size to ensure effective convergence.
    The parameters used were set as $\text{norm}=L_\infty$ and for Square attack, the $\text{n\_queries=5000}$.

\end{compactenum}

\paragraph{Concept Removal Interventions}
For concept removal interventions, we project embeddings into the nullspaces of classifiers. Here, the the rank $r$ corresponds to the dimensionality of the subspace identified and erased by the intervention, meaning that the number of dimensions removed is equal to the rank.\footnote{
    This is only true for binary properties $Z$ -- for variables that can take $n$ values with $n > 2$, the number of dimensions removed is $n \cdot r$.
} We experiment over the range of values $r \in [0, 1, ..., 40]$. 
We consider the following concept removal interventions:
\begin{compactenum}
    \item \textbf{INLP}
    We implement Iterative Nullspace
    Projection (INLP; \citealp{ravfogel2020inlp}) as follows: we train a series of classifiers $w_1, ..., w_n$, where in each iteration, embeddings are projected into the nullspace of the preceding classifiers $P_N(w_0) \cap \cdots \cap P_N(w_n)$.
    We then apply the combined projection matrix to calculate the final projection where $ P := P_N(w_1) \cap \dots \cap N(w_i) $, $X$ is the full set of embeddings, and $X_{\text{projected}} \leftarrow P(X)$.
    
    \item \textbf{RLACE}
    We implement Relaxed Linear
    Adversarial Concept Erasure (R-LACE; \cite{ravfogel2022adversarial}) which defines a linear minimax game to adversarially identify and remove a linear bias subspace.
    In this approach, \( \mathcal{P}_k \) is defined as the set of all \( D \times D \) orthogonal projection matrices that neutralize a rank $r$ subspace:
    $$P \in \mathcal{P}_k \leftrightarrow P = I_D - W^\top W {\vphantom{I_D - W^\top W}}$$
    The minimax equation is then solved to obtain the projection matrix $P$  which is used to calculate the final intervened embedding $X_{\text{projected}}$, similar to INLP 
    $$\text{min}_{\theta \in \Theta} \text{max}_{P \in \mathcal{P}_k} \sum_{n=1}^{N} \ell \left( y_n, g^{-1} \left( \theta^\top P x_n \right) \right)$$
    Hyperparameters for $P$ and $\theta$ were a learning rate of 0.005 and weight decay of 1e-5.
\end{compactenum}

\paragraph{\cfinlp}
We implement \cfinlp \cite{ravfogel2021counterfactual} by first running INLP, saving all classifiers, and using these to compute rowspace projections that push all embeddings to the positive $Z = \texttt{Pl}$ or negative $Z = \texttt{Sg}$ side of the separating hyperplane for all classifiers. That is, we compute

$$\hat{\rvh}^l_{Z=\texttt{Sg}} = P_N(\rvh) + \alpha \sum_{w \in \rmW}(-1)^{\textit{SIGN}(w \cdot \rvh)} (w \cdot \rvh) \rvh$$
$$\hat{\rvh}^l_{Z=\texttt{Pl}} = P_N(\rvh) + \alpha \sum_{w \in \rmW}(-1)^{1 - \textit{SIGN}(w \cdot \rvh)} (w \cdot \rvh) \rvh$$
where $P_N$ is the nullspace projection from INLP.

\section{Supplemental Results}\label{sec:all_supp_results}

\subsection{Final-Layer Completeness, Reliability, and Selectivity}\label{sec:interhyp}

\begin{figure*}[ht]
    \centering
    \begin{subfigure}[t]{0.48\textwidth}
        \centering
        \includegraphics[width=\linewidth]{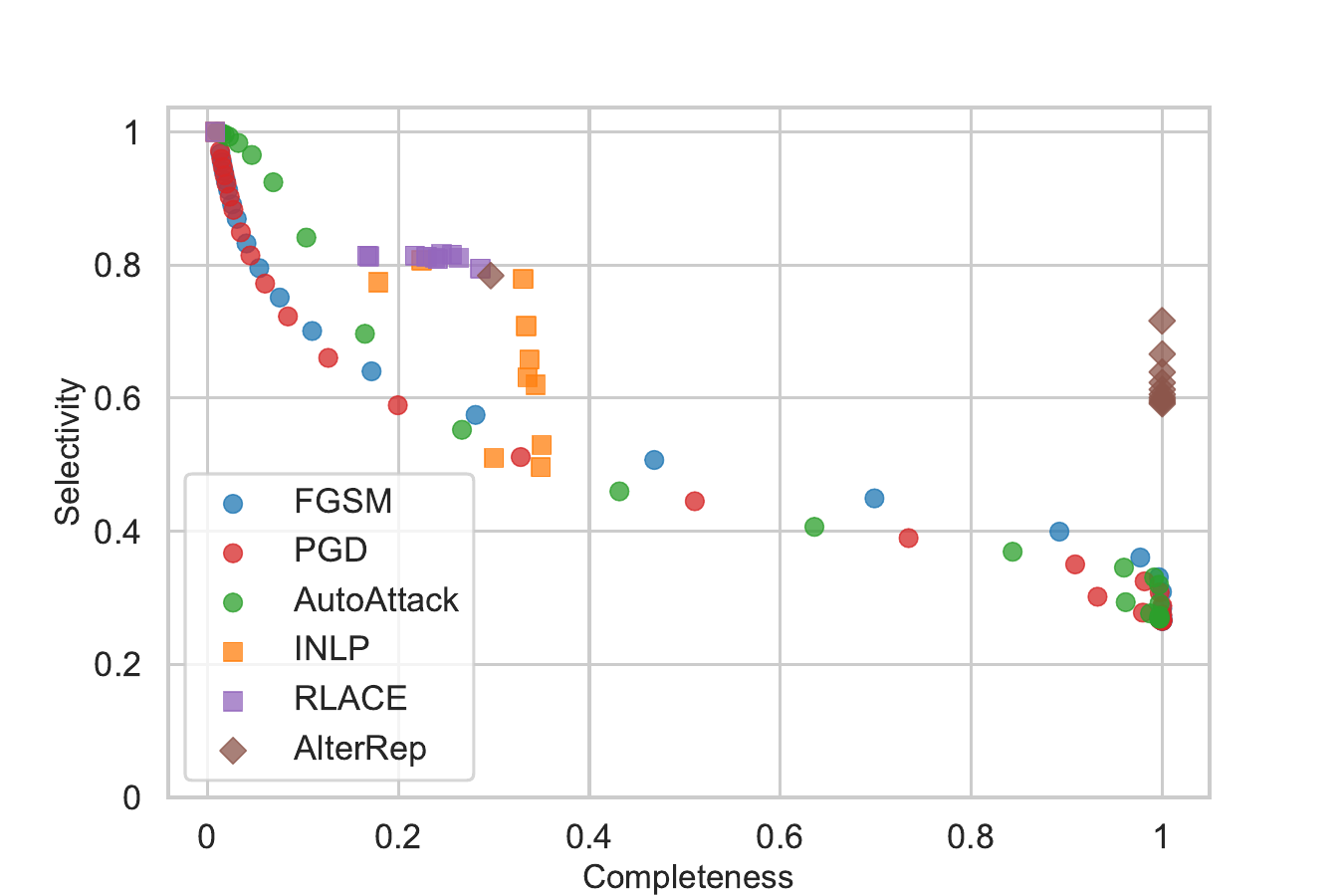}
        \caption{\textbf{Selectivity vs. Completeness}}
        \label{fig:comp_sel_bert}
    \end{subfigure}
    \hfill
    \begin{subfigure}[t]{0.48\textwidth}
        \centering
        \includegraphics[width=\linewidth]{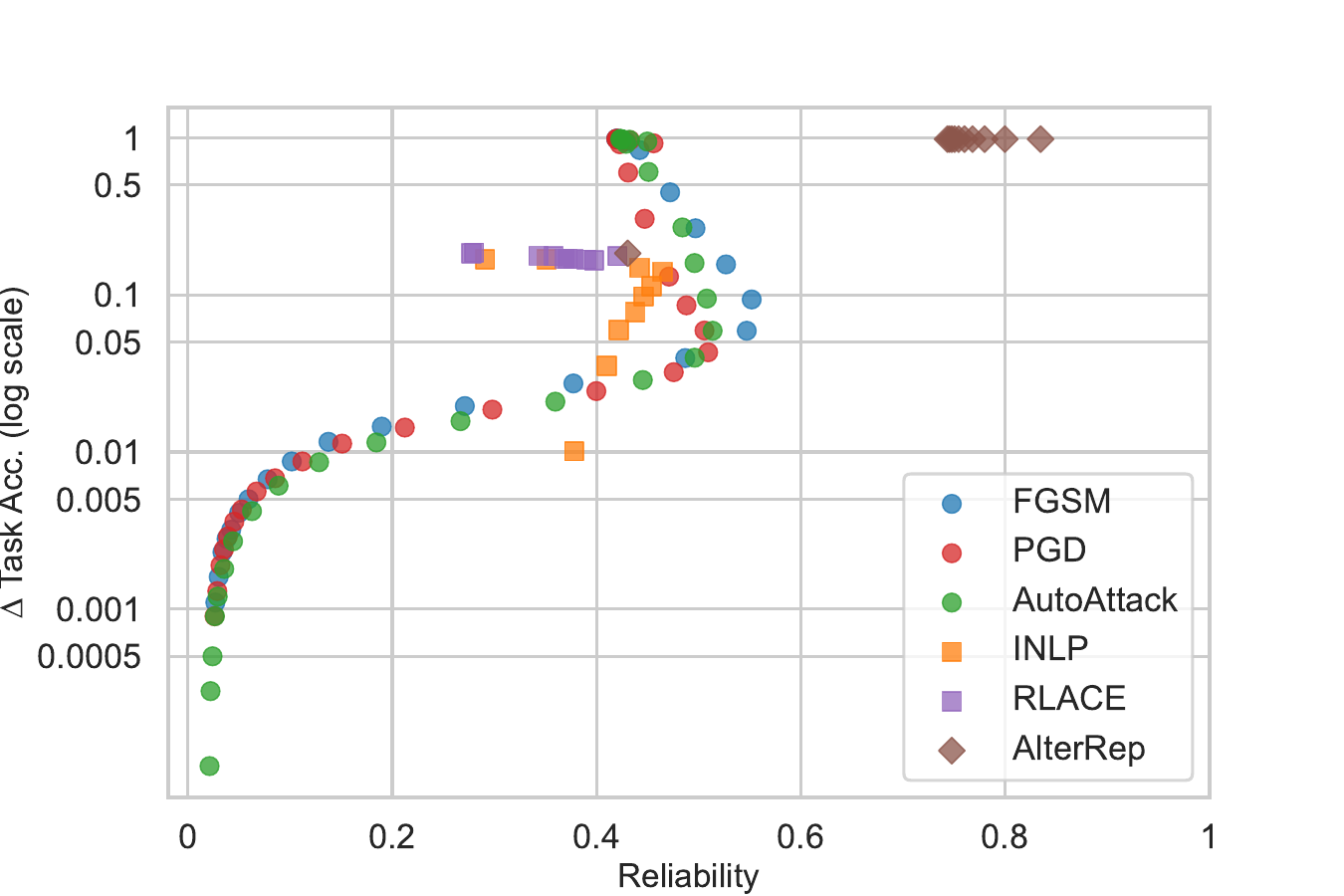}
        \caption{\textbf{$\Delta \text{Task Acc}$ vs. Reliability}}
        \label{fig:acc_rel_bert}
    \end{subfigure}
    \caption{\textbf{(BERT) Completeness, selectivity, reliability, and $\Delta \text{Task Acc}$} for all interventions in BERT's final layer. Each point in both plots corresponds to a different hyperparameter setting.}
    \label{fig:main_plots_bert}
\end{figure*}

\begin{figure*}[ht]
    \centering
    \begin{subfigure}[t]{0.48\textwidth}
        \centering
        \includegraphics[width=\linewidth]{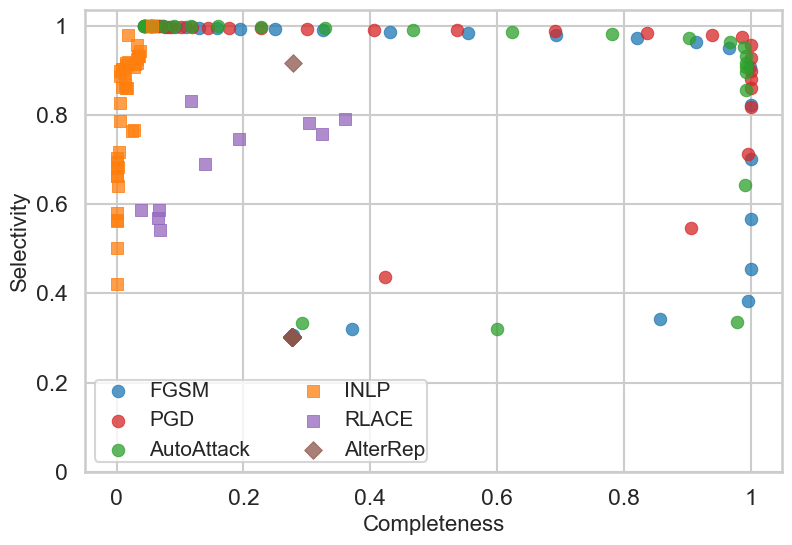}
        \caption{\textbf{Selectivity vs. Completeness}}
        \label{fig:comp_sel_gpt2}
    \end{subfigure}
    \hfill
    \begin{subfigure}[t]{0.48\textwidth}
        \centering
        \includegraphics[width=\linewidth]{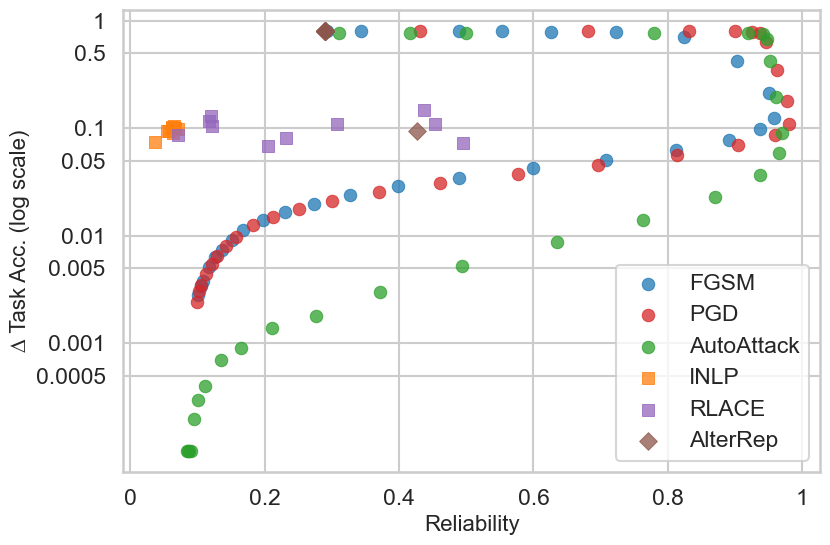}
        \caption{\textbf{$\Delta \text{Task Acc}$ vs. Reliability}}
        \label{fig:acc_rel_gpt2}
    \end{subfigure}
    \caption{\textbf{(GPT2) Completeness, selectivity, reliability, and $\Delta \text{Task Acc}$} for all interventions in the final layer of GPT2. Each point in both plots corresponds to a different hyperparameter setting.}
    \label{fig:main_plots_gpt2}
\end{figure*}

\begin{figure*}[ht]
    \centering
    \begin{subfigure}[t]{0.48\textwidth}
        \centering
        \includegraphics[width=\linewidth]{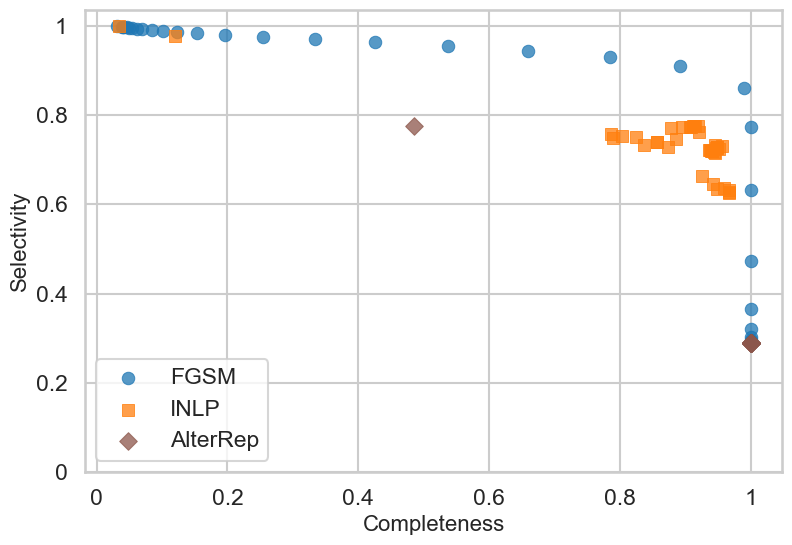}
        \caption{\textbf{(Pythia-1.4B) Selectivity vs. Completeness}}
        \label{fig:comp_sel_pythia14b}
    \end{subfigure}
    \hfill
    \begin{subfigure}[t]{0.48\textwidth}
        \centering
        \includegraphics[width=\linewidth]{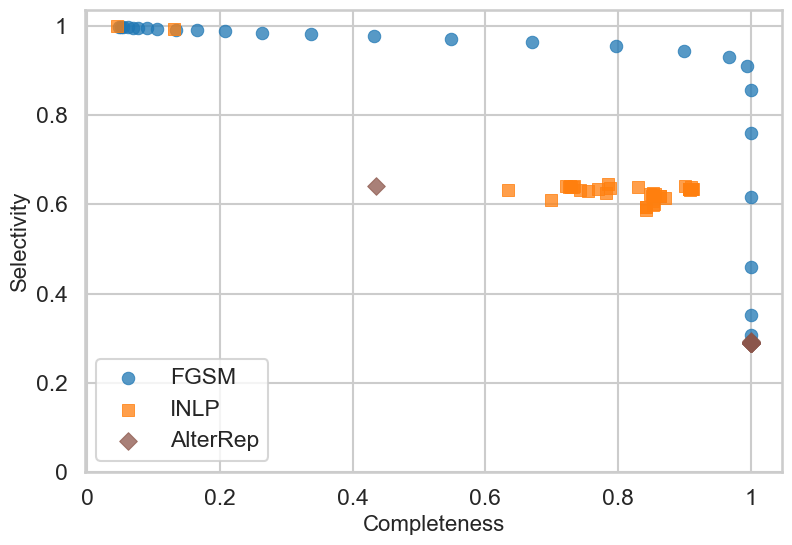}
        \caption{\textbf{(Pythia-6.9B) Selectivity vs. Completeness}}
        \label{fig:comp_sel_pythia69b}
    \end{subfigure}
    \hfill
    \begin{subfigure}[t]{0.48\textwidth}
        \centering
        \includegraphics[width=\linewidth]{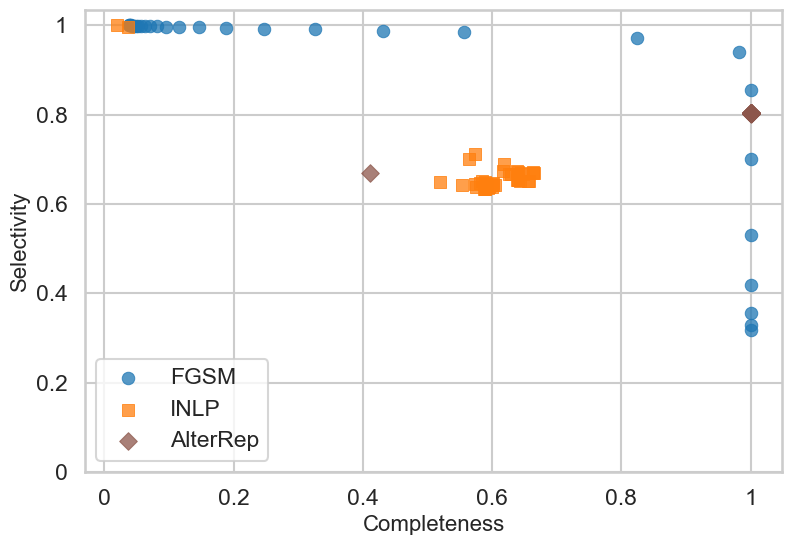}
        \caption{\label{fig:comp_sel_llama} \textbf{(Llama-3.2-3B-Instruct) Selectivity vs. Completeness}}
    \end{subfigure}
    \caption{\textbf{Completeness and selectivity} for all interventions in the final layer of \textbf{Pythia-1.4B, Pythia-6.9B, and Llama-3.2-3B-Instruct}. Each point in both plots corresponds to a different hyperparameter setting.}
    \label{fig:main_plots_pythia_bothlarge}
\end{figure*}

In \cref{fig:comp_sel_bert,fig:comp_sel_gpt2,fig:comp_sel_pythia14b,fig:comp_sel_pythia69b,fig:comp_sel_llama}, we visualize final-layer completeness and selectivity of intervention methods for all models except Pythia-160M, analogously to the \cref{fig:comp_sel_pythia} results reported in the main paper for Pythia-160M.
In \cref{fig:acc_rel_bert,fig:acc_rel_gpt2}, we show the relationship between $\Delta \text{Task Acc}$ and reliability for BERT and GPT2 (respectively), analogously to the \cref{fig:acc_rel_pythia} results reported in the main paper for Pythia-160M.

\clearpage

Additionally, in \cref{tab:alphaerr}, we report the standard error of completeness, selectivity, and reliability for BERT's maximum-reliability final-layer results displayed in \cref{tab:alpha}. Note that all scores have standard error $< 0.002$, and we observe the same pattern for all other models.

\begin{table*}
\footnotesize
\centering
\begin{tabular}{l|cccc}
\toprule
\textbf{} & ${C(\hat \rmH_{Z})}$ & $S(\hat \rmH_{Z})$ & $R(\hat \rmH_{Z})$ & $x_{opt}$\\
\midrule
INLP & $0.3308 \pm 0.0013$ & $0.7792 \pm 0.0012$ & $0.4644 \pm 0.0013$&$r=8$\\
RLACE & $0.2961 \pm 0.0013$ &$0.7782 \pm 0.0012$ &$ 0.4290 \pm 0.0014$&$r=33$\\ \midrule
\cfinlp & $1.0000 \pm 0.0000$ & $0.7162 \pm 0.0017$ & $\boldsymbol{0.8346} \pm 0.0012$& $\alpha=0.1$\\ \midrule\midrule
FGSM & $0.8923 \pm 0.0011$ &$0.3994 \pm 0.0018$& $0.5518\pm 0.0017$&$\varepsilon\!=\!0.112$ \\
PGD &  $0.7343 \pm 0.0016$& $0.3897 \pm 0.0018$& $0.5092 \pm 0.0016$&$\varepsilon\!=\!0.112$\\
AutoAttack & $0.8433 \pm 0.0013$ & $0.3692 \pm 0.0019$ & $0.5136 \pm 0.0018$&$\varepsilon\!=\!0.112$\\ \bottomrule
\end{tabular}
\vspace{0.5em}
\caption{\textbf{(BERT) Intervention scores for maximum-reliability hyperparameters} in the final layer, \textbf{with standard error included}. All scores are reported
for the hyperparameter $x_{opt}$ that maximizes the reliability of each respective method. Counterfactual methods are grouped above the double line, with concept removal methods below it.}
\label{tab:alphaerr}
\end{table*}

\begin{figure*}[htb!]
    \centering
    \begin{subfigure}{0.48\textwidth}
        \centering
        \includegraphics[scale=0.35,trim=2 2 2 2,clip]{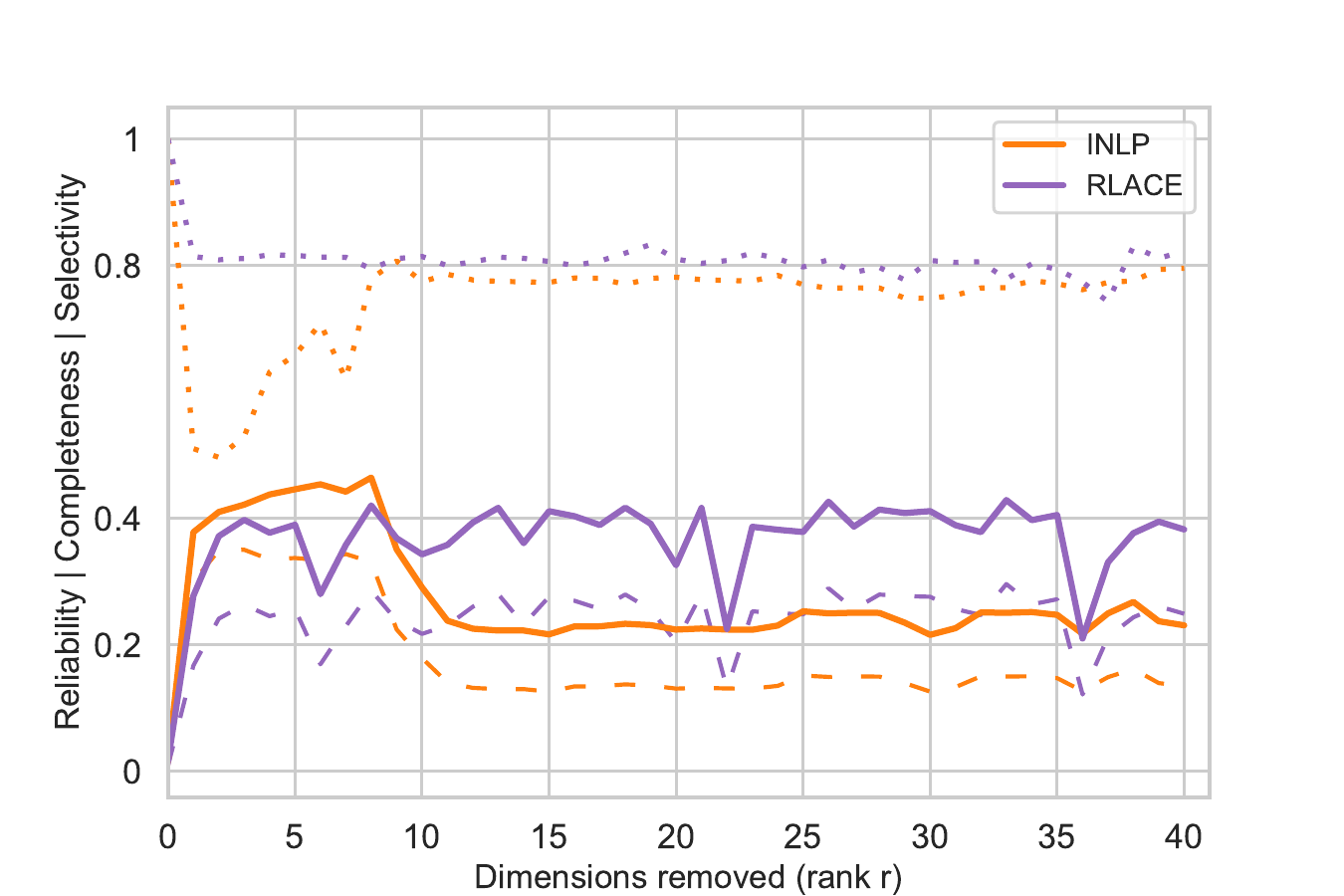}
        \caption{\textbf{Linear Removal (INLP, RLACE)}}
        \label{fig:inlpc_bert}
    \end{subfigure}
    \hfill
    \begin{subfigure}{0.48\textwidth}
        \centering
        \includegraphics[scale=0.35,trim=2 2 2 2,clip]{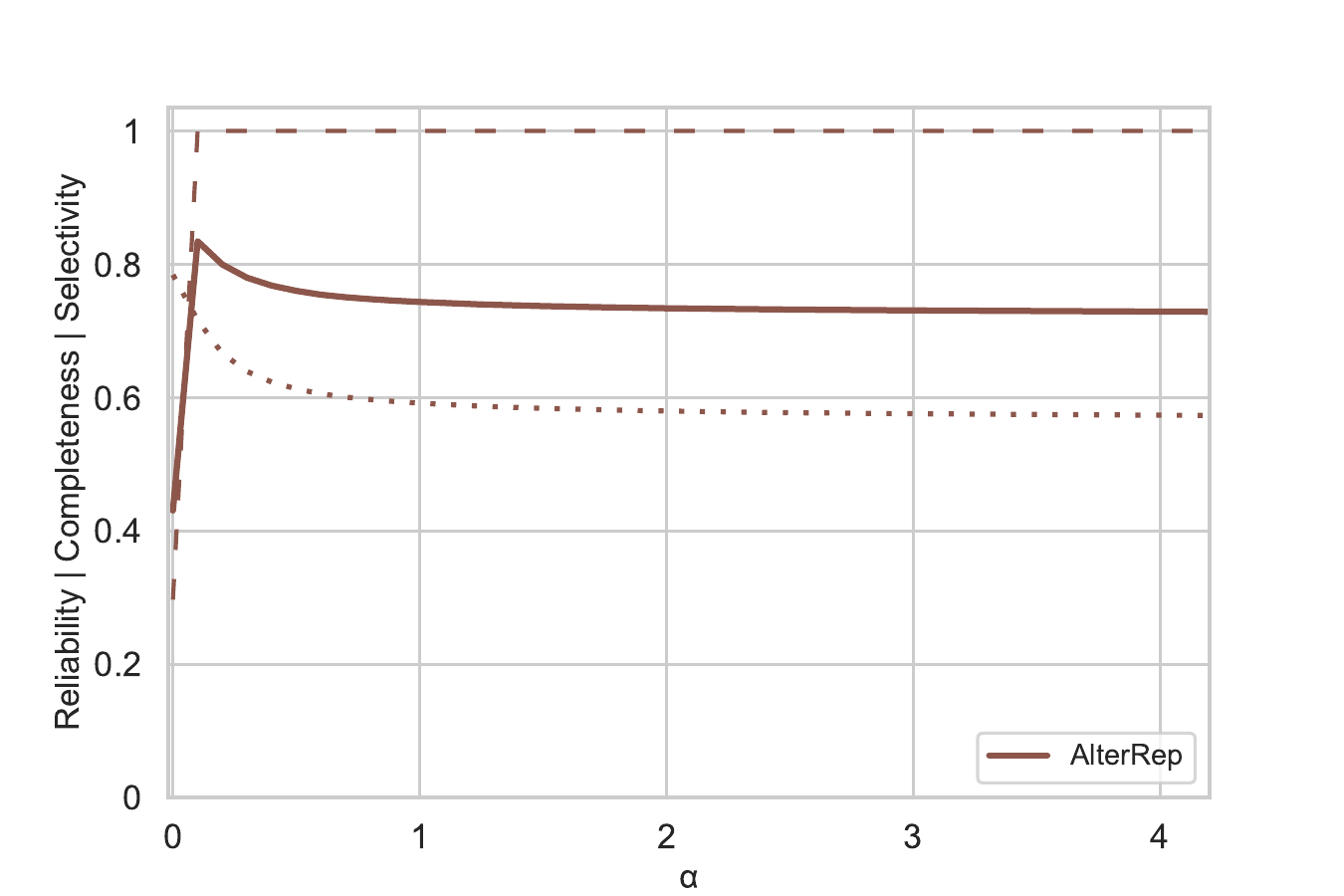}
        \caption{\textbf{Linear Counterfactual (\cfinlp)}}
        \label{fig:cfinlpc_bert}
    \end{subfigure}
    \hfill
    \begin{subfigure}{0.48\textwidth}
        \centering
        \includegraphics[scale=0.35,trim=2 2 2 2,clip]{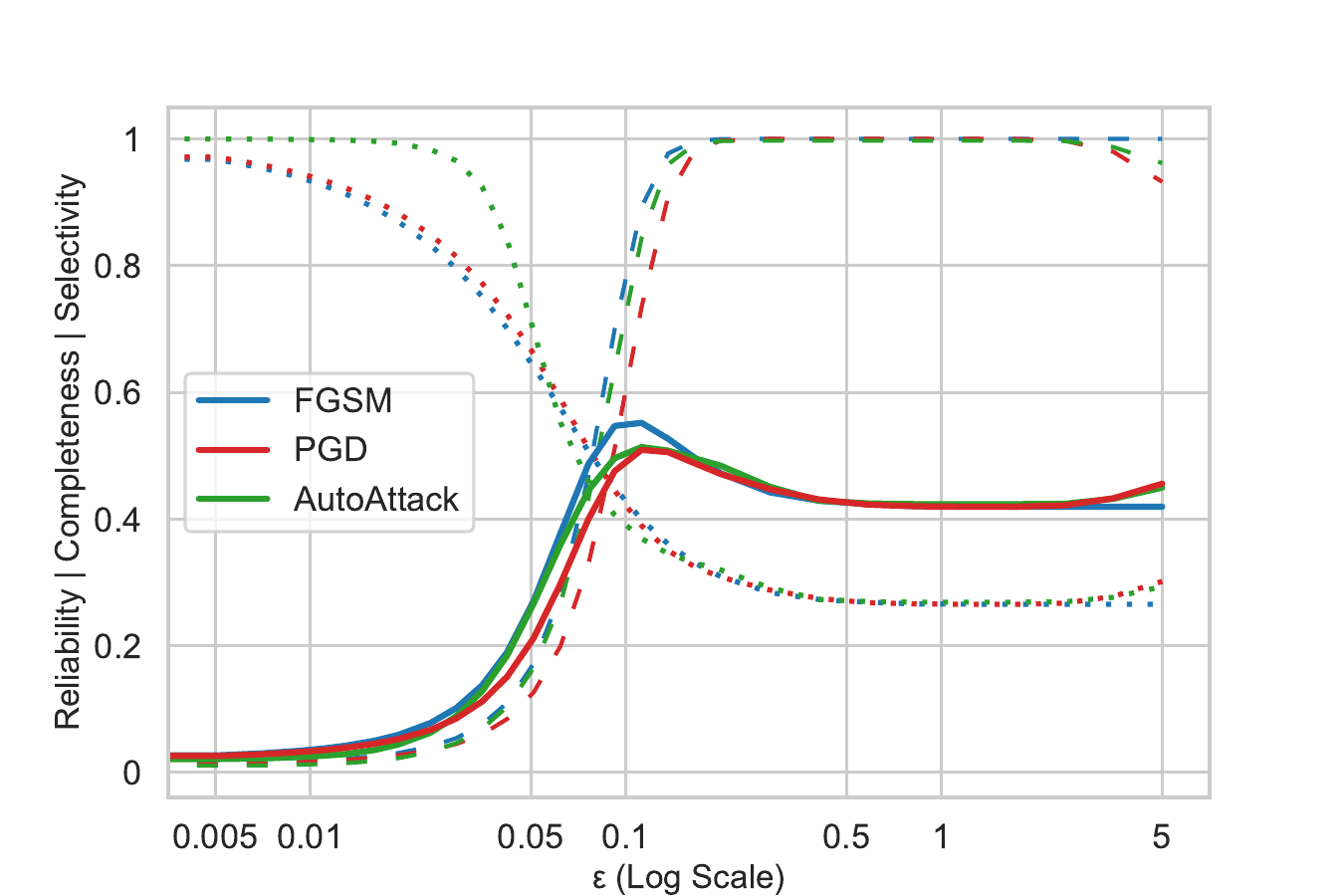}
        \caption{\textbf{Nonlinear Counterfactual (GBIs)}}
        \label{fig:cfgbic_bert}
    \end{subfigure}
    \caption{\textbf{(BERT) Reliability (solid), completeness (dashed), \& selectivity (dotted)} of all methods in BERT's final layer, by hyperparameter value.}
    \label{fig:hypers_bert}
\end{figure*}

\begin{figure*}[htb]
    \centering
    \begin{subfigure}[t]{0.48\textwidth}
        \centering
        \includegraphics[scale=0.35,trim=2 2 2 2,clip]{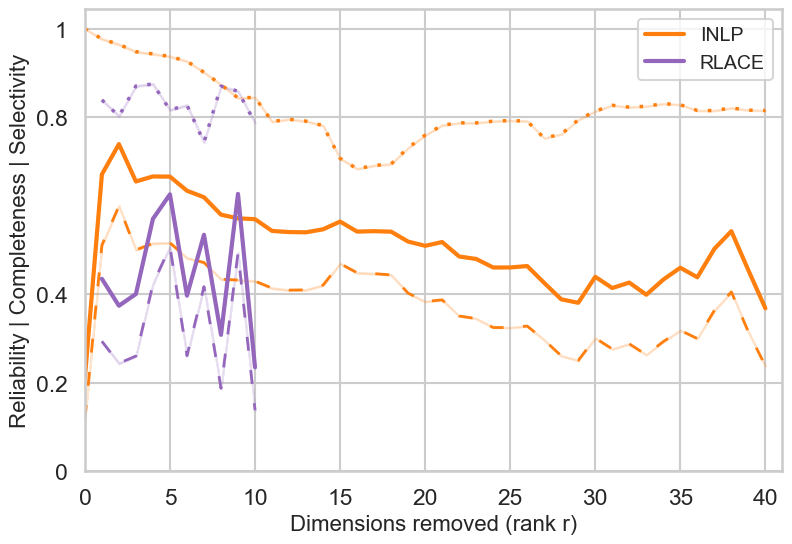}
        \caption{\textbf{Linear Removal (INLP, RLACE)}}
        \label{fig:inlpc_pythia}
    \end{subfigure}
    \hfill
    \begin{subfigure}[t]{0.48\textwidth}
        \centering
        \includegraphics[scale=0.35,trim=2 2 2 2,clip]{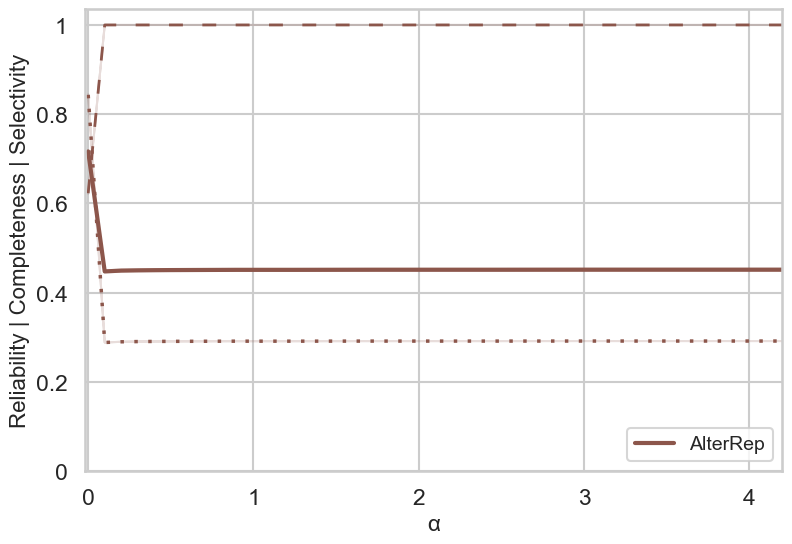}
        \caption{\textbf{Linear Counterfactual (\cfinlp)}}
        \label{fig:cfinlpc_pythia}
    \end{subfigure}
    \hfill
    \begin{subfigure}[t]{0.48\textwidth}
        \centering
        \includegraphics[scale=0.35,trim=2 2 2 2,clip]{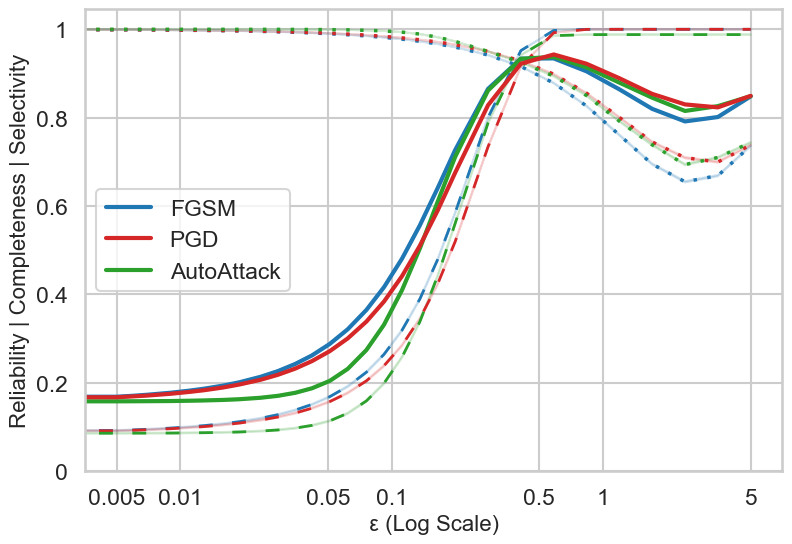}
        \caption{\textbf{Nonlinear Counterfactual (GBIs)}}
        \label{fig:cfgbic_pythia}
    \end{subfigure}
    \caption{\textbf{(Pythia-160M) Reliability (solid), completeness (dashed), \& selectivity (dotted)} of all methods in the final layer of Pythia-160M, by hyperparameter value.}
    \label{fig:hypers_pythia}
\end{figure*}

\paragraph{Hyperparameter Variation}\label{sec:hyper_variation}
In \cref{fig:hypers_bert,fig:hypers_pythia}, we observe that increasing the degree of control that interventions have over the representation of the target property by increasing the intervention hyperparameter associated with a given intervention type (i.e., $\varepsilon, \alpha$, or rank) generally leads to both improved completeness and decreased selectivity.

\clearpage

\subsection{Reliability by Layer}\label{sec:layerwise_other_models}
\begin{figure*}[ht]
    \centering
    \begin{subfigure}[t]{0.48\textwidth}
        \centering
        \includegraphics[width=0.95\columnwidth]{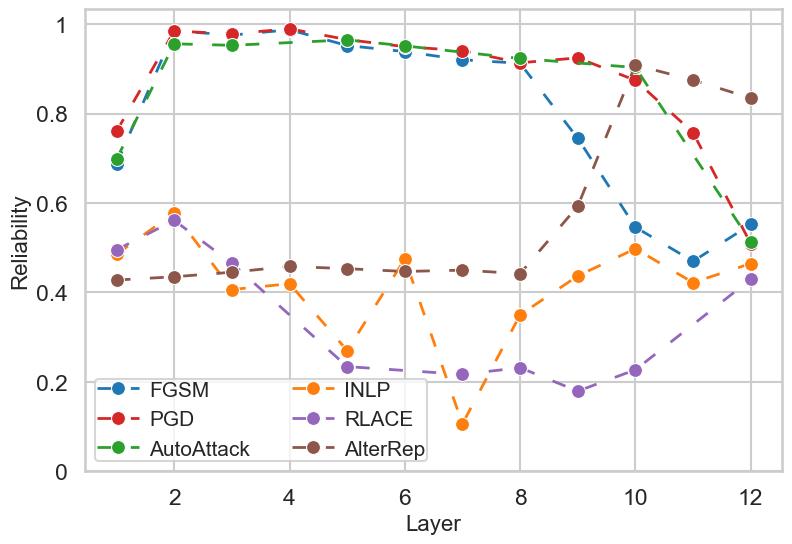}
        \caption{\textbf{BERT}}
        \label{fig:layerwise_bert}
    \end{subfigure}
    \hfill
    \begin{subfigure}[t]{0.48\textwidth}
        \centering
        \includegraphics[width=0.95\columnwidth]{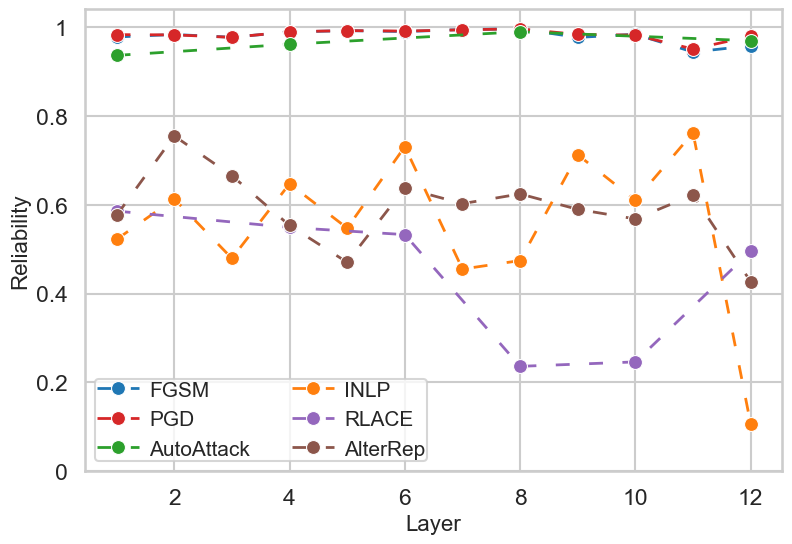}
        \caption{\textbf{GPT2}}
        \label{fig:layerwise_gpt2}
    \end{subfigure}
    \hfill
    \begin{subfigure}[t]{0.48\textwidth}
        \centering
        \includegraphics[width=0.95\columnwidth]{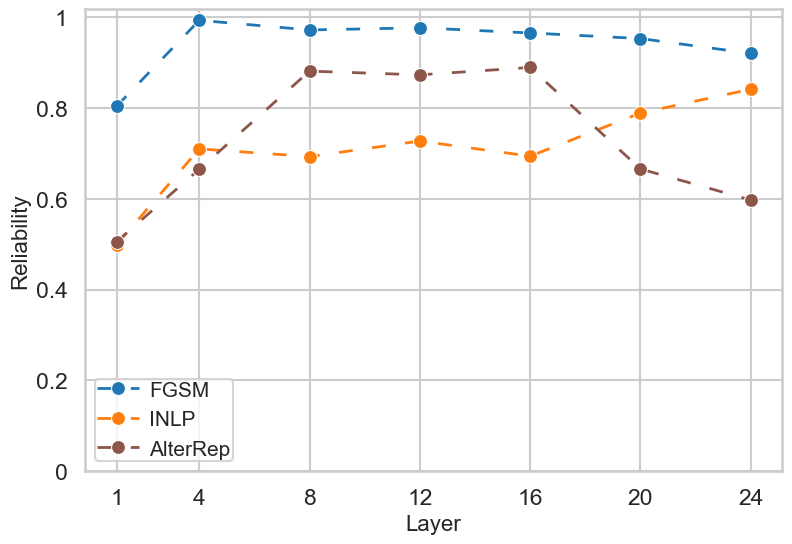}
        \caption{\textbf{Pythia-1.4B}}
        \label{fig:layerwise_pythia14}
    \end{subfigure}
    \hfill
    \begin{subfigure}[t]{0.48\textwidth}
        \centering
        \includegraphics[width=0.95\columnwidth]{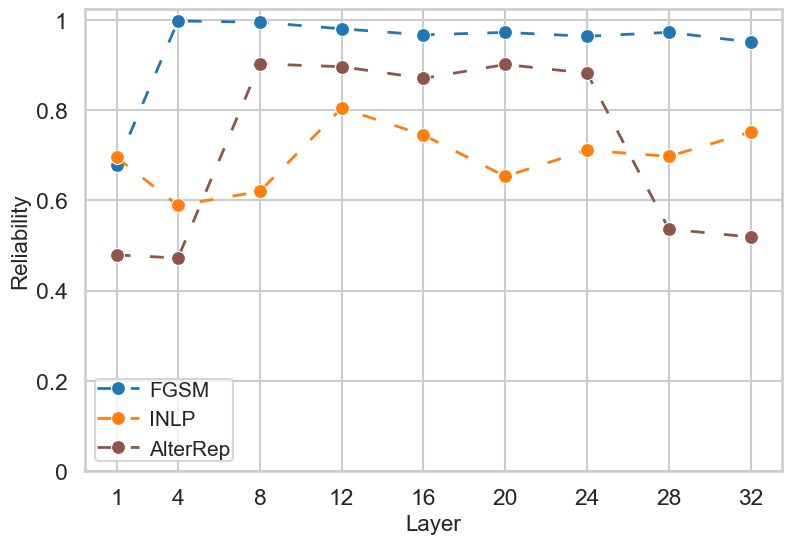}
        \caption{\textbf{Pythia-6.9B}}
        \label{fig:layerwise_pythia69}
    \end{subfigure}
    \begin{subfigure}[t]{0.48\textwidth}
        \centering
        \includegraphics[width=0.95\columnwidth]{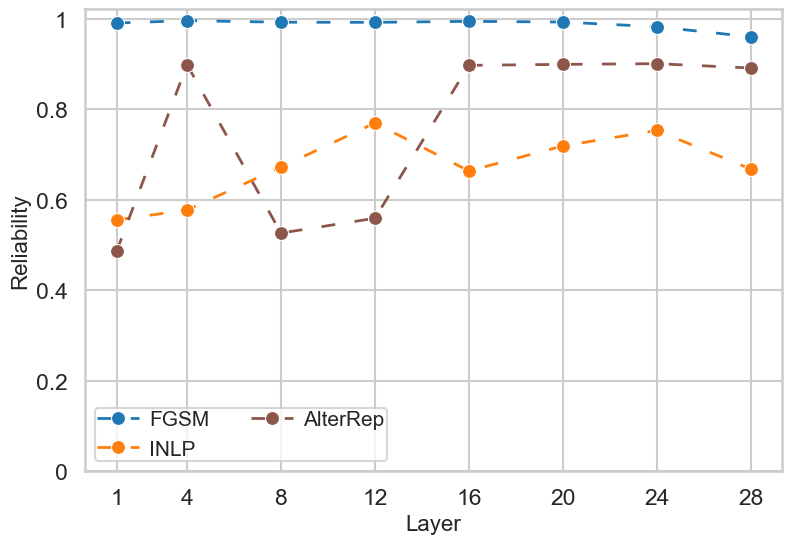}
        \caption{\textbf{Llama-3.2-3B-Instruct}}
        \label{fig:layerwise_llama}
    \end{subfigure}
\caption{\textbf{Maximum reliability by layer} across models for each intervention across all layers.}
\label{fig:layerwise_all_others}
\end{figure*}

In \cref{fig:layerwise_all_others}, we visualize maximum reliability of intervention methods across layers for all models (except Pythia-160M, which is reported in the main paper), analogously to the \cref{fig:layerwise_pythia} results reported in \cref{sec:earlierlayerresults}.

\clearpage

\subsection{Linear \Orcs}\label{sec:linorc}
\begin{figure*}[ht]
    \centering
    \begin{subfigure}[t]{0.48\textwidth}
        \centering
        \includegraphics[width=0.95\columnwidth]{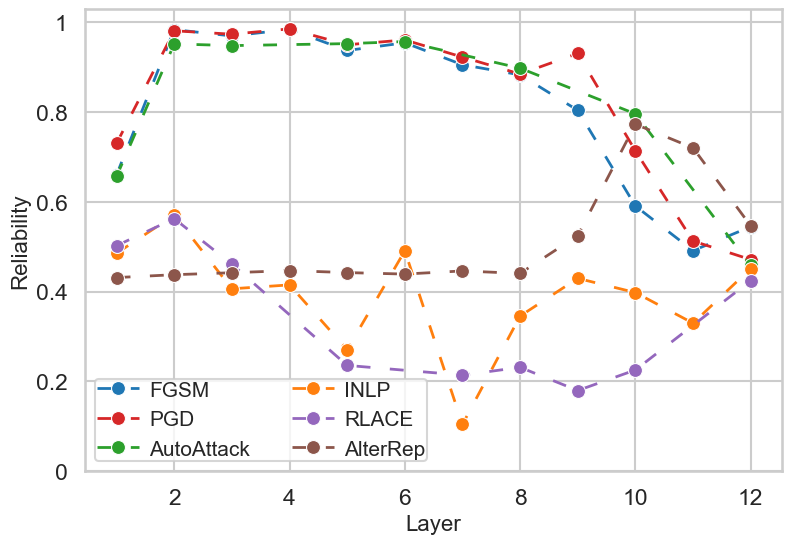}
        \caption{\textbf{BERT}}
        \label{fig:lin_layerwise_bert}
    \end{subfigure}
    \hfill
    \begin{subfigure}[t]{0.48\textwidth}
        \centering
        \includegraphics[width=0.95\columnwidth]{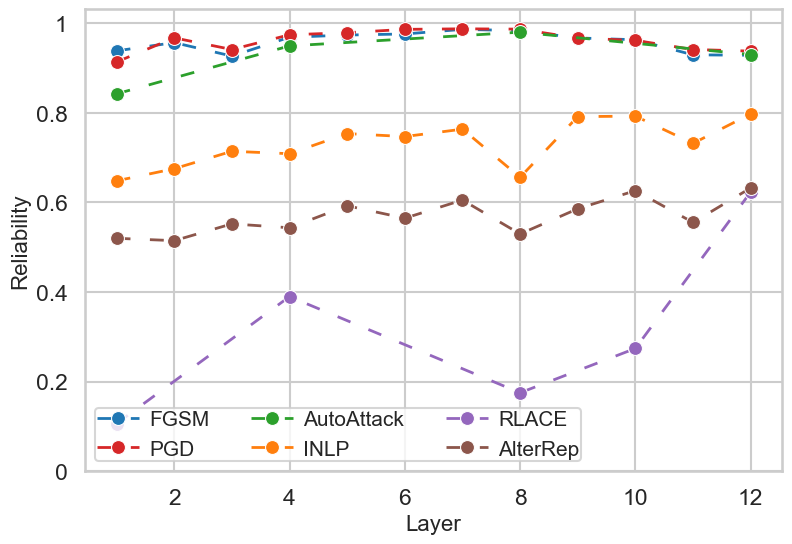}
        \caption{\textbf{Pythia-160M}}
        \label{fig:lin_layerwise_pythia}
    \end{subfigure}
    \caption{\textbf{Maximum reliability by layer} for each intervention across all layers, using \emph{linear \orcs}.}
    \label{fig:lin_layerwise_bert_pythia}
\end{figure*}

We present the reliability by layer for BERT and Pythia-160M using \emph{linear \orcs} in \cref{fig:lin_layerwise_bert,fig:lin_layerwise_pythia} (respectively). The main trends (specifically, reliability ordering of methods by layer) shown here are very similar to those using MLP \orcs, as shown in \cref{fig:layerwise_bert,fig:layerwise_pythia}, with the exception that the linear counterfactual method (\cfinlp) does not surpass the reliability of GBIs as strongly in the later layers (for BERT). 
The BERT result is not especially surprising, as linear \orcs are expected to be less resilient to linear interventions than MLP \orcs (as MLPs can also rely on nonlinearly-encoded information to make predictions) leading to lower selectivity and correspondingly lower reliability using linear interventions with linear \orcs compared to evaluations using nonlinear \orcs.
However, it is important to note that the overall ordering of methods, and the specific scores observed, are still remarkably similar between linear vs nonlinear \orcs for both models, indicating that the differences in reliability between linear and nonlinear methods are unlikely to be due to the (non)linearity of \orcs.

\clearpage

\subsection{\Orc Accuracy by Layer}\label{apx:layerwise_probe}
\begin{figure*}[ht]
    \centering
    \begin{subfigure}[t]{0.48\textwidth}
        \centering
        \includegraphics[width=0.95\columnwidth]{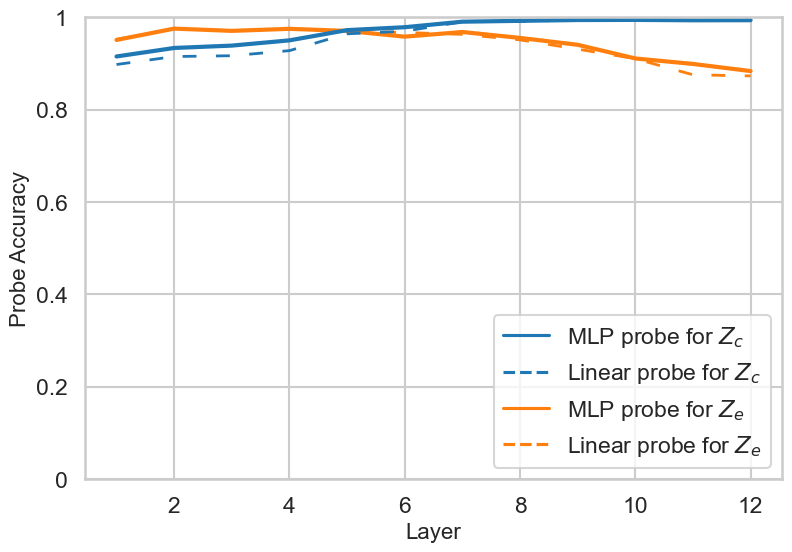}
        \caption{\textbf{BERT}}
    \end{subfigure}
    \hfill
    \begin{subfigure}[t]{0.48\textwidth}
        \centering
        \includegraphics[width=0.95\columnwidth]{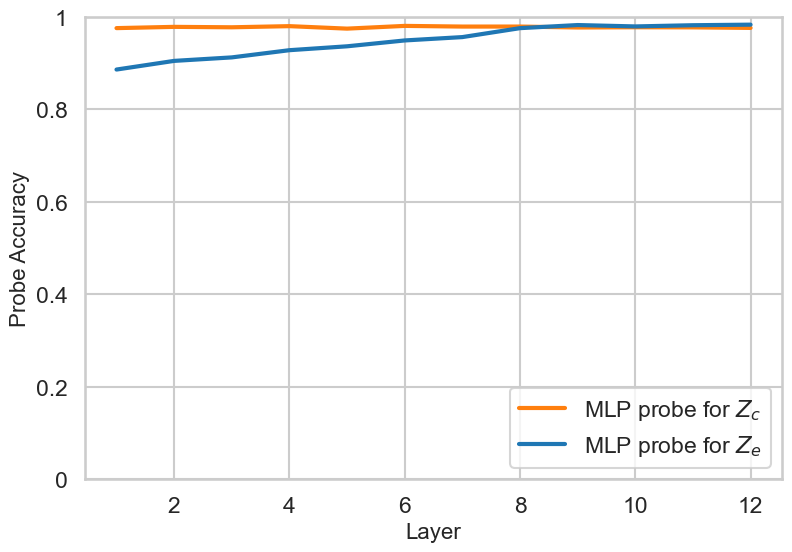}
        \caption{\textbf{GPT2}}
    \end{subfigure}
    \hfill
    \begin{subfigure}[t]{0.48\textwidth}
        \centering
        \includegraphics[width=0.95\columnwidth]{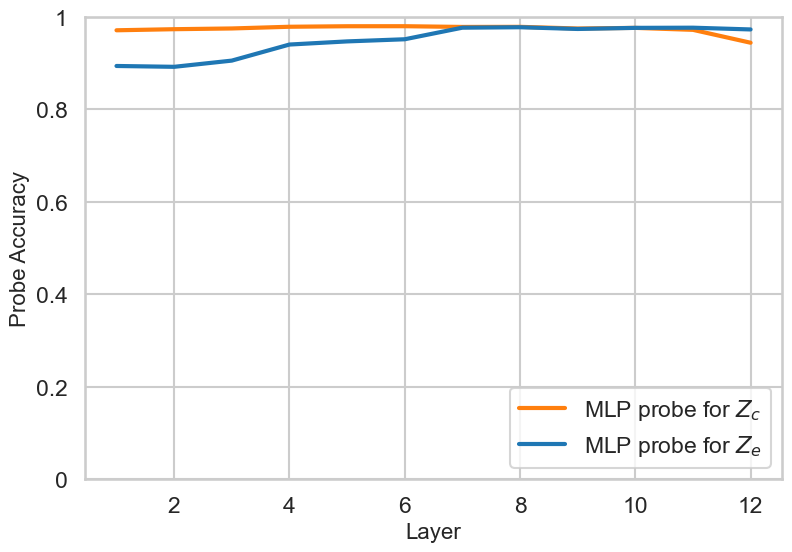}
        \caption{\textbf{Pythia-160M}}
    \end{subfigure}
    \hfill
    \begin{subfigure}[t]{0.48\textwidth}
        \centering
        \includegraphics[width=0.95\columnwidth]{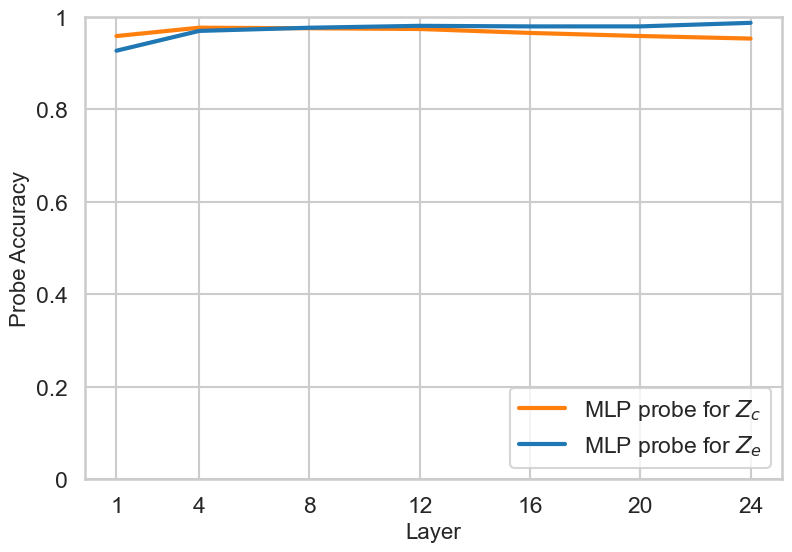}
        \caption{\textbf{Pythia-1.4B}}
    \end{subfigure}
    \hfill
    \begin{subfigure}[t]{0.48\textwidth}
        \centering
        \includegraphics[width=0.95\columnwidth]{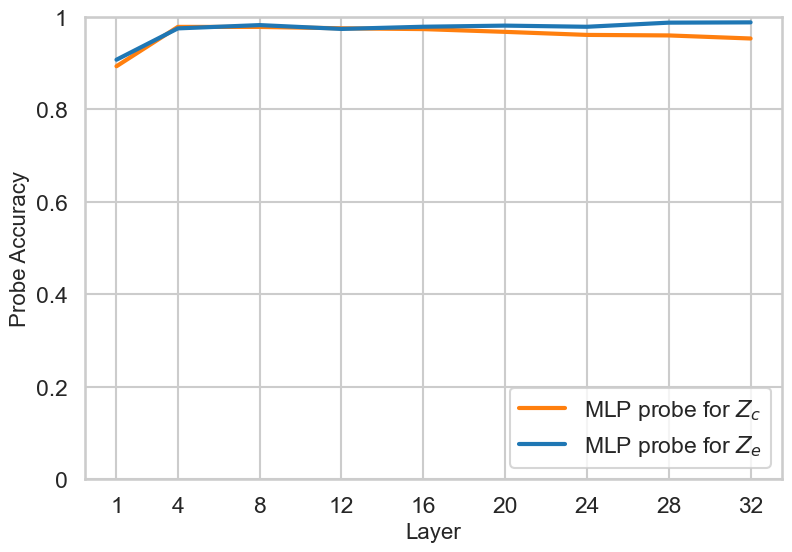}
        \caption{\textbf{Pythia-6.9B}}
    \end{subfigure}
    \hfill
    \begin{subfigure}[t]{0.48\textwidth}
        \centering
        \includegraphics[width=0.95\columnwidth]{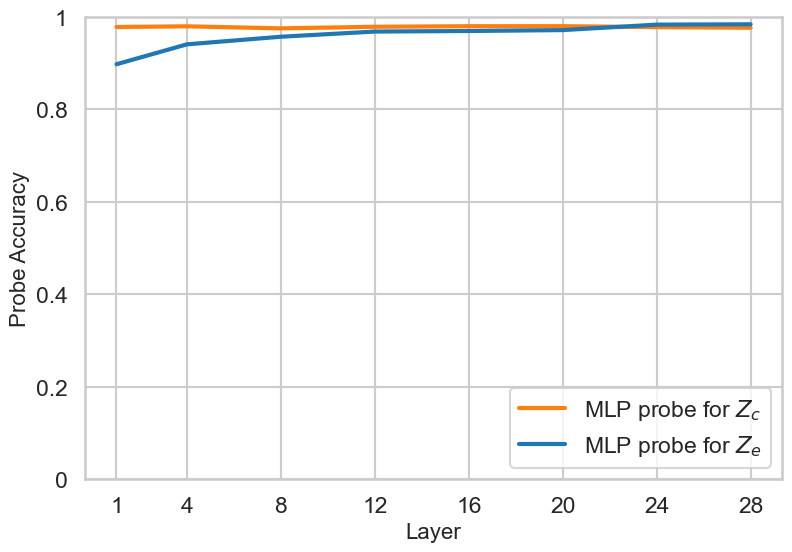}
        \caption{\textbf{Llama-3.2-3B-Instruct}}
    \end{subfigure}
    \caption{\textbf{\Orc Accuracy by Layer} across models.}
    \label{fig:layerwise_probe_accs}
\end{figure*}

In \cref{fig:layerwise_probe_accs}, we report the layerwise \orc accuracy across all models.

\clearpage

\subsection{Results on Indirect Object Identification (IOI) Task}\label{apx:ioi}
The Indirect Object Identification (IOI) task \cite{wang2023ioi} is a well-studied benchmark in mechanistic interpretability research, requiring models to identify the correct referent in sentences with multiple names. Each sentence has an initial dependent clause introducing two names (e.g., "After John and Mary went to the store..."), followed by a main clause where one person performs an action involving the other (e.g., "... John gave a bottle of milk to..."). The model must correctly complete the sentence with "Mary" (i.e., the indirect object) rather than repeating "John".

For our experiments, we define the causal variable $Z_c$ as denoting whether the first or second mentioned person in the sequence is the correct indirect object (each label has probability 0.5 in this dataset), and the environmental variable $Z_e$ as the tense of the root verb (e.g., "gives" vs "gave"), which is clearly irrelevant to solving the task.

\begin{figure*}[ht]
    \centering
    \begin{subfigure}[t]{0.48\textwidth}
        \centering
        \includegraphics[width=0.95\columnwidth]{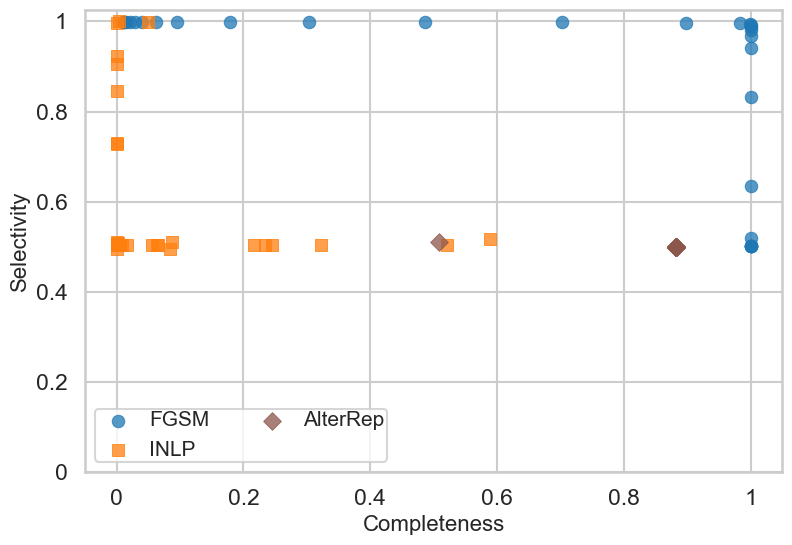}
        \caption{\textbf{Selectivity vs. Completeness}}
        \label{fig:ioi_selcomp}
    \end{subfigure}
    \hfill
    \begin{subfigure}[t]{0.48\textwidth}
        \centering
        \includegraphics[width=0.95\columnwidth]{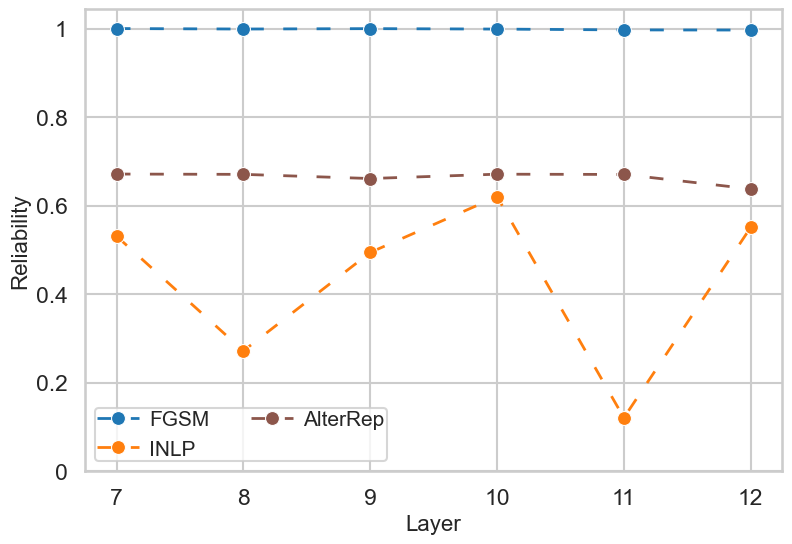}
        \caption{\textbf{Reliability by Layer}}
        \label{fig:ioi_layer}
    \end{subfigure}
    \caption{\textbf{Results on Indirect Object Identification (IOI) task.} Completeness, selectivity, and reliability for interventions using the IOI task and the GPT2 model.}
    \label{fig:ioi_gpt2}
\end{figure*}

Following \citet{zhang2024towards}, we study IOI in the context of GPT2.
Figure \ref{fig:ioi_gpt2} shows the relation between selectivity and completeness as well as the reliability of the various interventions across layers. (Note that we only display results on layers 7--12 because the earlier layers do not encode necessary information to predict the variables of interest -- i.e., in these layers, we cannot train a probe to predict the target features at sufficiently high accuracy, as also observed by \citealt{zhang2024towards}.)
The results are similar with the subject-verb agreement task on GPT2 in that FGSM (nonlinear) is more reliable than INLP and AlterRep (linear) at all layers; but on the IOI task, we find that AlterRep is more reliable than INLP at all layers (for subject-verb agreement, these methods' relative performances varied on GPT2).

\end{document}